\documentclass{article}
\usepackage{microtype}
\usepackage{graphicx}
\usepackage{subcaption}
\usepackage{booktabs}
\usepackage{hyperref}

\usepackage[main, final]{neurips_2026}

\usepackage{amsmath}
\usepackage{amssymb}
\usepackage{mathtools}
\usepackage{amsthm}
\usepackage{hyperref}
\usepackage{url}
\usepackage{enumitem} 
\usepackage{booktabs}
\usepackage{multirow}
\usepackage{makecell}   
\usepackage{xspace}
\usepackage{graphicx}
\usepackage{wrapfig}
\usepackage{subcaption}
\usepackage{caption}
\usepackage{tcolorbox}
\usepackage{fancyvrb}
\usepackage{fvextra}
\usepackage{algorithm}
\usepackage{algorithmic}
\usepackage{listings}
\usepackage[dvipsnames]{xcolor}
\definecolor{myred}{RGB}{200, 50, 50}
\definecolor{mygreen}{RGB}{50, 180, 50}

\newcommand{\method}{\textsc{IPG}\xspace}
\usepackage{booktabs,makecell,multirow}
\newcommand{\bx}{\mathbf{x}}
\newcommand{\bh}{\mathbf{h}}

\newcommand{\bW}{\mathbf{W}}

\DefineVerbatimEnvironment{DiffVerb}{Verbatim}{
    commandchars=!\{\},, 
    fontsize=\small,
    breaklines=true,     
    frame=none,           
    breaksymbol=,             
    breaksymbolsep=0pt,       
    breaksymbolindentleft=0pt       
}
\usepackage[capitalize,noabbrev]{cleveref}

\theoremstyle{plain}

\theoremstyle{definition}

\theoremstyle{remark}

\usepackage[textsize=tiny]{todonotes}
\usepackage[utf8]{inputenc} 
\usepackage[T1]{fontenc}    
\usepackage{hyperref}       
\usepackage{url}            
\usepackage{booktabs}       
\usepackage{amsfonts}       
\usepackage{nicefrac}       
\usepackage{microtype}      
\usepackage{xcolor}         
\newlength{\affwidth}
\newcommand{\aff}[1]{\makebox[\affwidth][c]{#1}}

\title{From Outcome to Representation: Tracing Reasoning Mechanisms  through Integrated Policy Gradient}
\author{%
  Changming Li \\
  ShanghaiTech University \\
  \And
  Kaixing Zhang \\
  ShanghaiTech University \\
  \And
  Yingdong Shi \\
  ShanghaiTech University \\
  \And
  Haoyun Xu \\
  ShanghaiTech University \\
  \And
  Zheng Zhang \\
  ShanghaiTech University \\
  \And
  Kaitao Song \\
  \aff{Apple} \\
  \And
  Kan Ren\thanks{Corresponding author: \texttt{renkan@shanghaitech.edu.cn}} \\
  ShanghaiTech University \\
}

\begin{document}

\maketitle

\begin{abstract}
Large Language Models (LLMs) demonstrate remarkable reasoning capabilities, yet the
internal mechanisms driving these multi-step processes remain opaque. Existing mechanistic interpretability approaches often rely on superficial text-pattern co-occurrences or capture only short-term effects, struggling to trace the long-horizon, outcome-driven influence of internal components such as neurons or sparse features in multi-step reasoning.
In this paper,  we introduce Integrated Policy Gradient (\method), a novel, training-free framework that brings policy-based attribution into mechanistic interpretability for LLM reasoning.
Grounded in \textit{outcome-oriented} and \textit{sequential-influence-aware} principles, 
it backpropagates outcome-level, non-differentiable behavioral signals (e.g., reasoning outcome  correctness) through entire inference trajectories, utilizing path integration to isolate outcome-relevant mediators at component level for reliable attribution.
This provides a  linear approximation of Causal Mediation Analysis (CMA) in reasoning. 
Empirical evaluations demonstrate that \method achieves precise localization and reveals mechanistic insights by identifying sparse, concentrated subsets of internal components associated with specific reasoning behaviors.
Our method also enables modulation of reasoning capabilities, offering a powerful and reliable framework for understanding LLM reasoning.  
The project page is at \url{https://foundation-model-research.github.io/IPG}.
\end{abstract}

\section{Introduction}

Large Language Models (LLMs) have shown outstanding performance in addressing natural language processing (NLP) tasks~\citep{touvron2023llama,browngpt32020,team2024qwq}.
Beyond simple next-token prediction, modern LLMs now demonstrate sophisticated reasoning abilities, including structured, step-by-step problem solving~\citep{weichainofthought2022,khot2023decomposedpromptingmodularapproach}.
Despite these advancements, a central question remains unanswered:\textit{ which internal components are critical to multi-step reasoning, and how do their contributions across reasoning steps shape the final task outcome.}
This lack of transparency poses challenges for truthfulness, safety, and controllability in reasoning systems~\citep{wang2025comprehensive}.

Mechanistic interpretability studies, which characterize internal computations responsible for model behaviors,
suggest that the internal representation space of LLMs encodes reasoning-related concepts~\citep{hjer2025improving,galichin2025icoveredbaseshere,wang2026llm}. 
Existing interpretability approaches for LLM reasoning fall into three categories.
The first analyzes individual internal components, such as neurons~\citep{rai-yao-2024-investigation} or sparse 
features~\citep{galichin2025icoveredbaseshere,fang2026controllable}, applying mediation analysis~\citep{stolfo2023a} on \emph{short-term effect}, e.g., arithmetic, or on correlations between their activations and text patterns, e.g., reasoning-related tokens.
The second comprises conventional attribution methods, including Integrated Gradients (IG), NeuronSHAP, and Layer-wise Relevance Propagation (LRP)~\citep{sundararajan2017axiomatic,neushapnips2020,attenlrpicml2024}, which quantify the contributions of internal components to model predictions through gradient integration, Shapley-value estimation, or relevance propagation.
The third derives control vectors to induce desired behaviors~\citep{hjer2025improving,wang-etal-2025-beyond-prompt}, but they rely heavily on well-designed contrastive \emph{input} pairs, 
making it difficult to guarantee that the intended target states are genuinely elicited~\citep{hjer2025improving}.
These prevalent approaches face one or both of the following limitations:
(i) \textit{Reliance on co-occurrences}: 
they rely solely on input prompts or activation co-occurrences with text-pattern, failing to isolate the model's inner components like neurons or sparse features~\citep{huben2024sparse}
that genuinely drive reasoning behaviors.
(ii) \textit{Neglecting sequential influence}: they capture only short-term and local effects, e.g., arithmetic, instead of how sequential influence cumulatively contributes to the final outcome (e.g., task correctness).

\begin{figure*}[t!]
    \centering
    \includegraphics[width=1.0\textwidth]{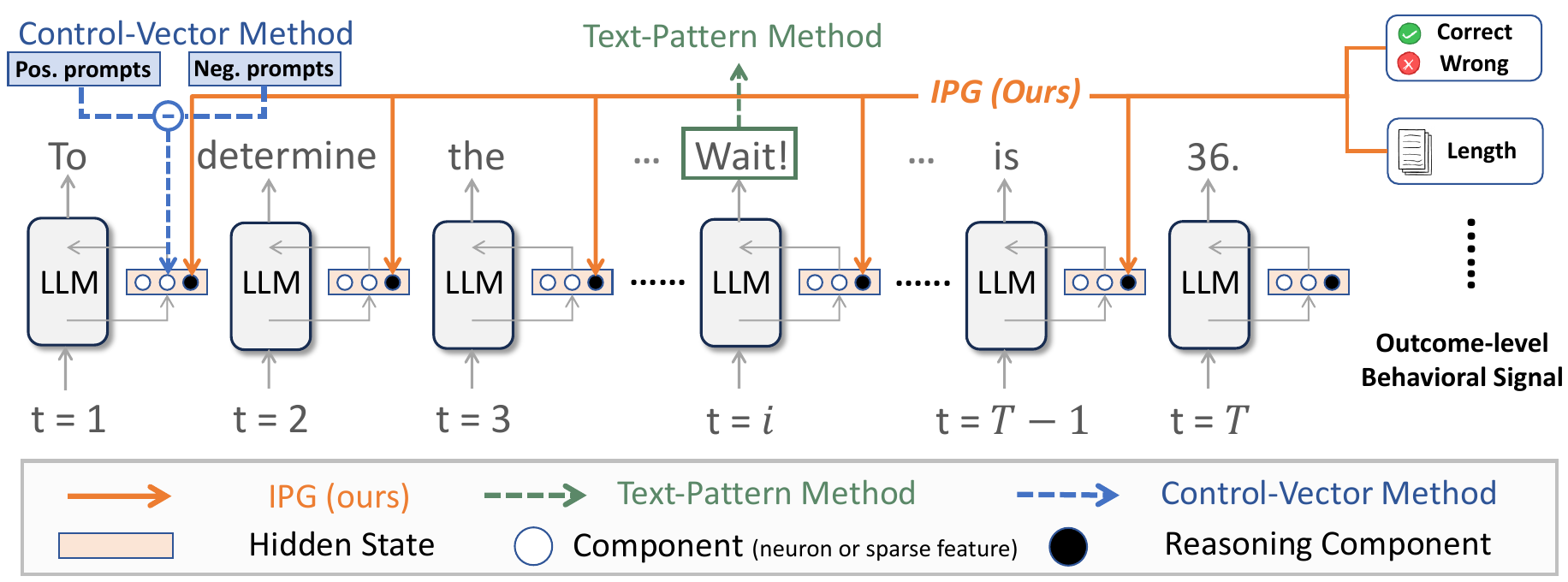}
    \caption{Interpretability paradigms for reasoning in LLMs, regarding
(i) our proposed Integrated Policy Gradient method that attributes outcome-level behavioral signals.
$t$ is the reasoning time step;
(ii) text-pattern methods that associate high activations with special tokens (e.g., ``Wait'');  and
(iii) control-vector methods that rely on human-craft contrastive input pairs.}\label{fig:pipeline}
\vspace{-20pt}
\end{figure*}

\begin{wrapfigure}{r}{0.5\textwidth}
    \centering
    \vspace{-10pt} 
    \includegraphics[width=\linewidth]{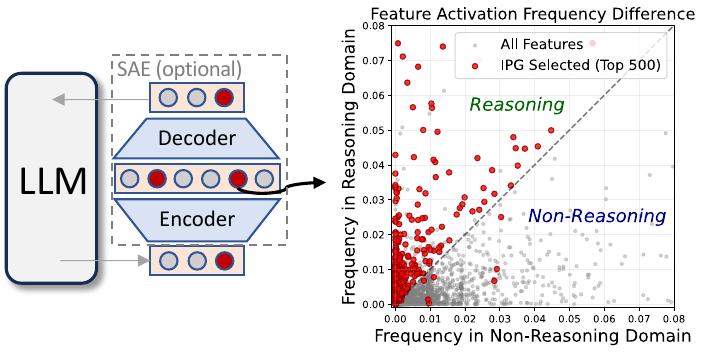}
    \vspace{-20pt}
    \caption{SAE feature activation frequency across reasoning and non-reasoning domains with \method-selected features (red) and all features (gray).
    }\label{fig:activation_analysis}
    \vspace{-12pt}
\end{wrapfigure}
To bridge this gap, we propose Integrated Policy Gradient (\method), a novel training-free framework for attributing and interpreting reasoning in LLMs by identifying outcome-relevant components.
\method incorporates policy gradient~\citep{mnih2016asynchronous,schulman2017proximal} to back-propagate outcome-based 
behavioral
signals through the entire inference trajectory to the model's internal components. 
This enables attribution of cumulative, long-range effects, overcoming challenges of sparse and non-differentiable outcome signals.
\method further accumulates policy gradients along the path from a baseline (e.g. zero value) to the original activation of each component for reliability in attribution~\citep{sundararajan2017axiomatic}.
We compare our approach with existing interpretability paradigms in \cref{fig:pipeline}.
Additionally,~\cref{fig:activation_analysis} highlights the specific reasoning components isolated by \method.

From a causal perspective, \method can be viewed as a tractable approximation to Causal Mediation Analysis (CMA)~\citep{pearl2001direct}, which quantifies the indirect effects of internal components (mediators) on model outcomes through counterfactual interventions.
Directly estimating such effects requires repeated counterfactual evaluations, which becomes computationally prohibitive for long reasoning trajectories.
By approximating these mediation effects without explicit counterfactual rollouts, \method enables scalable component-level attribution of reasoning.

We evaluate and validate the components identified by \method on representative open-source LLMs and its effectiveness in interpreting LLM reasoning.
These components support improvements in reasoning
performance and exhibit greater approximation on causal mediation than the competing attribution methods.
Our effectiveness also transfers across datasets and domains.
Beyond these  evaluations, our spatial, semantic,
and temporal analyses provide complementary views of the
identified components: where they are concentrated, what
reasoning-related patterns (e.g., logical planning or calculation) they are associated with, and when
they become active during generation.

In summary, our main contributions are threefold, centered on improving the mechanistic understanding and controllability of reasoning in LLMs. 
(i) We introduce Integrated Policy Gradient (\method), a novel training-free framework towards reasoning mechanism that is both \textit{sequential-influence-aware} and \textit{outcome-oriented}, attributing reasoning behaviors directly to internal components based on long-horizon outcomes.
To the best of our knowledge, we are the first to bring policy-based attribution 
into mechanistic interpretability for LLM.
(ii) We demonstrate that \method provides mechanistic insights into how multi‑step reasoning is internally organized. 
(iii) We show that \method provides reliable and predictable modulation of key reasoning behaviors.

\section{Related work}
\label{relate_work}

\textbf{Reasoning in Large Language Models. }
Reasoning has become a core capability of Large Language Models (LLMs). 
Existing improvements in LLM's reasoning mainly include prompt-elicited \citep{weichainofthought2022,khot2023decomposedpromptingmodularapproach} and training-induced reasoning~\citep{guo2025deepseek,openai2024o1,team2024qwq}.
Training-induced methods involve reinforcement learning (RL) to optimize reasoning abilities with outcome reward signals. 
Alternatively, other methods utilize supervised fine-tuning, where models are explicitly trained on datasets of reasoning chains \citep{yue2023mammothbuildingmathgeneralist} or distilled from more powerful reasoning models \citep{guo2025deepseek}. Despite these improvements, the internal representations and encoding of reasoning in these models remain unclear.

\textbf{Mechanistic Interpretability in Language Models for Reasoning.}
Mechanistic interpretability (MI) seeks to reverse-engineer the computations of language models by examining their internal representations~\citep{bereska2024mechanistic}. Reasoning, however, requires multi-step and long-horizon internal computation, posing unique challenges for the prevalent MI methods~\citep{reasoningMI}.
Conventional attribution methods, including Integrated
Gradients~\citep{sundararajan2017axiomatic},
Neuron Shapley~\citep{neushapnips2020}, and
LRP~\citep{lrp2016,attenlrpicml2024}, identify components relevant
to a single step, but do not explicitly address outcome-based
credit assignment across generated trajectories.
Reasoning-specific studies target reasoning-related concepts for internal components interpretation, including arithmetic~\citep{rai-yao-2024-investigation} neurons and reasoning-specific  features~\citep{galichin2025icoveredbaseshere,fang2026controllable} using Sparse Autoencoders~\citep{gao2025scaling,huben2024sparse}, focusing on correlating activations with surface text patterns (e.g., reasoning tokens like ``wait'').
Other work aims to control model behavior by adding vectors derived from contrastive input pairs~\citep{hjer2025improving,he2026reasoning,sheng2025on,ma2026sparse}. 
These methods require careful annotations and prompt design, which struggle to ensure the intended target states are truly elicited~\citep{hjer2025improving}.
In addition, ~\citep{stolfo2023a} applies CMA to identify causal modules in reasoning, but they primarily focus on short-term tasks such as simple arithmetic.
Overall, existing methods rely on co-occurrence correlations and fail to capture the cumulative, multi-step influences inherent in the reasoning process. 
In contrast, our approach is \textit{sequential-influence-aware} and \textit{outcome-oriented}, capable of attributing reasoning behavior without relying on human annotations or hand-crafted text patterns.

\section{Methodology}

\subsection{Preliminary: Reasoning in Large Language Models}
\textbf{Reasoning Task Formulation.} 
Let $F$ be a pretrained large language model. Formally, we define LLM reasoning as a sequential generation process. Given an input query $\mathbf{x}$, the language model generates a reasoning trajectory $\tau = (a_1, a_2, \dots,a_t,\ldots, a_T)$
of length $T$ autoregressively, consisting of generated tokens $a_t$, and produces a final answer $y$. We denote the overall reasoning process as $(\tau,y)=F(\mathbf{x})$. 
This characterizes reasoning as 
a \emph{multi-step} decision-making process with long-term sequential dependency,
where each token-generation time step can be associated with various reasoning behaviors.

\textbf{Reasoning Component Space.} 
LLM internal representations encode concepts related to reasoning behaviors~\citep{galichin2025icoveredbaseshere,wang2026llm}.  
Our framework localizes these concepts at either the neuron-level (hidden states $z_i$) or the disentangled feature-level components ($f_i$) with Sparse Autoencoders (SAEs)~\citep{huben2024sparse}.

Let $\mathbf{z}^l_t=[z_{t,1}^l,\ldots,z_{t,n}^l] \in \mathbb{R}^n$ denote the hidden state in layer $l$ at generation step $t$ under the input query $\mathbf{x}$, and $\mathbf{f}^l=[f_{t,1}^l,\ldots,f_{t,m}^l] \in \mathbb{R}^m$ its corresponding SAE feature representation (details in Appendix~\ref{Sec:apptrainsae}).
Together, these representations constitute our \emph{reasoning component space}.
We define a unified variable $\mathbf{h}^l_t \in \mathbb{R}^d$ to represent the state under analysis, where $\mathbf{h}_t^l= [h_{t,1}^l,\ldots,h_{t,d}^l]\in \{\mathbf{z}^l_t, \mathbf{f}^l_t\}$, with $d=n$ for neuron-level and $d=m$ for feature-level.

Building on this formulation, we decompose the reasoning process into \emph{trajectory-level behaviors} and their underlying \emph{component-level mechanisms}.
Concretely, at each generation step $t$ along the trajectory $\tau$, the model instantiates an internal representation $\mathbf{h}^{l}_t$. 
This representation mediates the generation dynamics, governing both the immediate next-token prediction and the accumulation of long-horizon dependencies.
Formally, we express the generation of the reasoning trajectory $\tau$ and the final response $y$ as a function of the input $\mathbf{x}$ conditioned on the sequence of internal states. For simplicity, we omit the notation $l$ to use $\bh_t$ to represent target component in the subsequent part.
\begin{equation}
    (\tau,y)=F(\mathbf{x}\mid\mathbf{h}^l_{1:T})\; .
\end{equation}
Crucially, the formulation reveals that reasoning outcomes emerge from the cumulative contribution of internal components across the full trajectory, a property we term \emph{sequential-influence}. 
It considers how a component affects the multi-step decisions that lead to the final outcome $(\tau, y)$.

\textbf{Outcome-oriented Measurement of Reasoning Behavior.} Reasoning behavior is the model's underlying mechanism during the reasoning process (i.e., how reasoning unfolds along the multi-step trajectory), including reasoning capability \citep{mondorf2024beyond} and reasoning strength \citep{sheng2025on}. 
Based on the outcome-oriented principle, we define a signal function $J(\cdot)$ to further help assess the model's internal components regarding various reasoning behaviors.
Specifically, 
$J$ can be instantiated as a binary reward $J(\tau)\in \{0, 1\}$ on outcome to measure reasoning capability, or as the total number of generated tokens $J(\tau)=|\tau|$ to quantify reasoning strength, where $\tau$ denotes a reasoning trajectory.
Notably, the definition of $J(\cdot)$ is highly \textit{flexible}, as it can be readily extended to capture more specific reasoning behaviors, such as reflection, as well as domains other than reasoning, including safety. Further discussion is provided in Appendix~\ref{Sec:appotheraspects}.

\subsection{Integrated Policy Gradient (IPG) Method}
\label{Sec:ipgattribution}

Existing interpretability methods in LLM reasoning fall short due to three critical limitations:
(i) Reliance on co-occurrences between activations and predefined text-patterns~\citep{rai-yao-2024-investigation,galichin2025icoveredbaseshere}
(ii) failure to capture \emph{sequential-influence} characterized by cumulative outcomes~\citep{hjer2025improving};
(iii) exact CMA is intractable for long sequences, and standard gradient-based linear approximations~\citep{cmasurveyma2026,dai-etal-2022-knowledge} struggle with sparse, non-differentiable feedback (e.g., final-answer correctness).

To address these challenges, we now introduce the Integrated Policy Gradient (\method) (in \cref{fig:pipeline}), a framework that proceeds by extending policy gradients to the representation space, refining the attribution via path integration, and promoting robustness through dataset-level aggregation.

\textbf{Policy Gradients in Representation Space.} 
We integrate policy-gradient methods~\citep{mnih2016asynchronous,schulman2017proximal}, which propagate outcome-aware signals (e.g., correctness rewards) backward through the trajectory. 
To obtain  attribution of long-horizon reasoning behavior,
we extend the perspective of policy gradient~\citep{mnih2016asynchronous,schulman2017proximal} from parameter space to representation space by propagating outcome-based signals back to internal components $\mathbf{h}_t$ (see Appendix~\ref{Sec:appjustificationpg} for details). 
Concretely, for a signal function $J(\cdot)$ that assesses reasoning behavior, with $\bh$ as the variable of interest, the policy gradient in representation space is
\begin{equation}
\small
\label{eq:policygrad}
\frac{\partial J}{\partial h_i}
=\;
\mathbb{E}_{\tau\sim\pi_{\theta}}\!\left[\sum_{t=1}^{T}
\frac{\partial}{\partial h_{t,i}}\log\pi_{\theta}\big(a_t\mid s_t;\mathbf{h}_t\big)
\cdot A^{\pi}(s_t,a_t)\right],
\end{equation}
where $s_t = (\mathbf{x}, a_1, \dots, a_{t-1})$ represents the contextual state at time step $t$ (consisting of the initial query $\mathbf{x}$ and all preceding prefix tokens)
and $a_t$ is the next token at the time step $t$.
$\tau=(a_1,\dots,a_T)$ is a reasoning trajectory following $\pi_\theta$ which is the policy parameterized by $\theta$ in the LLM $F$ that maps from $s_t$ to $a_t$.
$A^{\pi}(s_t,a_t)$ is the advantage function estimating the benefit of taking $a_t$ under $s_t$. 
Empirically, this value is estimated based on the feedback derived from the signal function $J(\cdot)$.

Notably, rather than using policy gradients to update model parameters $\theta$, we compute gradients with respect to \(\mathbf{h}\) to obtain attribution signals that reflect the long-horizon impact of internal activations on reasoning outcomes.
These signals form the basis for quantifying their effects on reasoning behavior and selecting components for subsequent causal interventions.

\textbf{IPG Attribution. } Raw gradients can be noisy, capturing only local sensitivity at a specific point.
To address this, we propose Integrated Policy Gradient (\method), guided by our  \textit{outcome-oriented}, \textit{sequential-influence-aware} principle, it quantifies the  mediation effect of an internal component on the final reasoning outcome.
The integral accounts for the global contribution of a component by aggregating its effect across the entire interpolation path from a baseline to its original value.

Specifically,
to quantify the influence of each component $h_{t,i}$ (the mediator) at step $t$ on a reasoning-behavior measure $J$, we define the attribution score (details in Appendix~\ref{app:ig_justification}) as
\begin{equation}
\small
\label{eq:ipg}
\begin{aligned}
\mathrm{IPG}(i;\bx)
&=
(h_i-h'_i)
\int_{0}^{1}
\frac{\partial J}{\partial h_i(\alpha)}
\, d\alpha
\\
&=
\sum_{t=1}^{T} (h_{t,i}-h'_{t,i})
\int_{0}^{1}
\mathbb{E}_{\tau\sim\pi_{\theta}}
\left[
\frac{\partial}{\partial h_{t,i}}
\log \pi_{\theta}
\!\left(
a_t \mid s_t;\,
\mathbf{h}'_{t}+\alpha(\mathbf{h}_{t}-\mathbf{h}'_{t})
\right)
\cdot A^{\pi}(s_t,a_t)
\right]
\, d\alpha .
\end{aligned}
\end{equation}
where $\mathbf{h}'_{t}$ is the relative baseline value of $\mathbf{h}_t$ at step $t$, and 
$\mathbf{h}_{t}(\alpha)=\mathbf{h}'_{t}+\alpha(\mathbf{h}_{t}-\mathbf{h}'_{t})$ represents the intervened hidden state along the path.
Deriving from~\cref{eq:policygrad},
for each generated trajectory \(\tau=(a_1,\dots,a_T)\),
the log-probability of selecting token $a_t$ under the interpolated state is weighted by the advantage function $A^{\pi}(s_t,a_t)$.
This step allows us to backpropagate non-differentiable final reasoning outcomes (e.g., binary  correctness) directly to intermediate internal components.
We accumulate these policy gradients along the path from the baseline $\mathbf{h}'$ to the active state $\mathbf{h}$~\citep{sundararajan2017axiomatic,dhamdhere2018how},
capturing the global mediation effect rather than just local variance.
\method bypasses the limitations of exact CMA and standard linear approximations. It reduces gradient noise and produces  robust signals for causal intervention.
We provide ablation in Appendix~\ref{Sec:appablapg}.

To ensure robustness, the \method attribution score is computed per sample and aggregated across a dataset.  
Let $\mathcal{D}=\{\bx^{(k)}\}_{k=1}^{M}$ be a  small supporting dataset of $M$ reasoning questions.
For each component index $i$ and each sample index $k$, we compute the per-sample attribution score $\mathrm{IPG}\big(i;{\bx^{(k)}}\big)$ (\cref{eq:ipg}).  
These scores are aggregated into a global importance statistic, for example, the mean attribution score $S_i$ for component $h_i$, and the top-$p$
components $\mathbf{P} =\{i_1,\dots,i_p\}$ are treated as the most influential components 
\begin{equation}
\small
\label{eq:ipgscore}
S_i \;=\; \frac{1}{M}\sum_{k=1}^{M} \mathrm{IPG}\!\big(i;\bx^{(k)}\big), 
\quad 
\mathbf{P} \;=\; \operatorname*{arg\,top-p}_{i \in [d]}\; S_i \;. 
\end{equation}

\textbf{Control Reasoning Behavior via \method .}
\label{Sec:intervene}
We intervene on the components indexed by $i \in \mathbf{P}$ identified in~\cref{eq:ipgscore} in the model, following interpretability practices~\citep{dai-etal-2022-knowledge}.
Formally, for each component $h_{t,i}$ at step $t$, we apply a multiplicative scaling factor as
\begin{equation}
h_{t,i}\leftarrow\text{Intervene}(h_{t,i})= \gamma h_{t,i}\;
\label{Eq:intervene}
\end{equation}
By setting different values of $\gamma$, we can enhance ($\gamma > 1$) the identified components, thereby controlling  the reasoning behavior in the target model.

Finally, at each step $t$, we substitute the original internal state $\mathbf{h_t}$ with the modified state vector $\tilde{\mathbf{h_t}}$ that is composed of the intervened components $\tilde{h}_{i}$ and original components $h_{j \notin \mathbf{P}}$. 
This yields a modified generation process, producing a new trajectory $\tilde{\tau}$ and response $\tilde{y}$:
\begin{equation}
(\tilde{\tau}, \tilde{y}) = F(\mathbf{x} \mid \tilde{\mathbf{h}}_{1:T}).
\label{eq:intervened_generation}
\end{equation}
\cref{fig:activation_analysis} visualizes this effect, and we empirically verify in~\cref{sec:reasoning_identicontrol} that the modified trajectory reflects the targeted modulation of reasoning behavior.

\textbf{Discussion: Flexibility of \method.} 
Importantly, the \method framework is \emph{data-efficient}. The attribution process only needs to be performed \textit{once} on a frozen model. 
After the set of components $\mathbf{P}$ is identified, interventions can be applied repeatedly during inference without any need for recomputing attributions. Furthermore, our empirical results confirm that \method can accurately pinpoint the core components even with a relatively small dataset $\mathcal{D}$ (Appendix~\ref{Sec:appattrhyper}). 
Besides, the mediator $\bh$ is not restricted, but can represent any interpretable module within the network architecture. 
Finally, the framework can be seamlessly integrated with advanced policy gradient algorithms, such as GRPO~\citep{shao2024deepseekmath}, to further refine the estimation of reasoning outcomes (Appendix~\ref{Sec:appalgvariants}).

\section{Experiments}
\label{sec:experiments}

In this section, we empirically evaluate Integrated Policy Gradient (\method) for  attributing and interpreting reasoning behaviors in Large Language Models (LLMs).
Our evaluation is driven by three core research questions (\textbf{RQs}) designed to progressively validate our approach.
\textbf{RQ1 (Control Efficacy):} Can \method identify internal components that exert a verifiable, sequential influence on reasoning behaviors?
\textbf{RQ2 (Generalizability):} Are the identified reasoning components consistent and transferable across different benchmarks and domains?
\textbf{RQ3 (Mechanistic Insights):} What deeper, fine-grained insights into the internal reasoning processes of LLMs does \method reveal?
\vspace{-8pt}
\subsection{Experimental Setup}
\label{Sec:expsetup}
\textbf{Evaluation Benchmarks. }
For comprehensive evaluation, we choose several reasoning-related benchmarks, including GSM8K~\citep{cobbe2021trainingverifierssolvemath}, MATH500~\citep{hendrycks2021measuringmathematicalproblemsolving}, AIME2024~\citep{aime2024} and GPQA-Diamond~\citep{rein2023gpqagraduatelevelgoogleproofqa}.

\textbf{Target Models. }
We experiment with two representative open-source LLMs: Qwen2.5-Math-1.5B-Instruct~\citep{yang2024qwen25mathtechnicalreportmathematical} and Llama3.1-8B-Instruct~\citep{grattafiori2024llama3herdmodels}, and additionally evaluate DeepSeek-Qwen-1.5B~\citep{guo2025deepseek}.  
1

\textbf{Baselines. }
We compare our \method and \method-SAE (integrated with k-SAE~\citep{gao2025scaling}) against several baselines: \textit{Random} (RND.), which samples neurons uniformly, and \textit{Activation} (ACT.), which ranks by activation magnitude.  
We also include \textit{Reasoning Neuron} (R.N.)~\citep{rai-yao-2024-investigation}, an SAE-based method (SAE-R)~\citep{galichin2025icoveredbaseshere} and \textit{Control Vector} (C.V.)~\citep{hjer2025improving},  which represent interpretable approaches using text patterns and contrastive data pairs for reasoning-related interventions.
More details are in Appendix~\ref{Sec:baselines}.

\textbf{Evaluation settings.}
Following prior work \citep{cheng2024smallagentrockempowering,tang2025unlockinggenerallongchainofthought}, we consistently employ a deterministic \textit{greedy} decoding strategy during inference to remove variance and ensure reproducible mechanistic analysis and also provide half-width 95\% CI for \method. 
To demonstrate that our method is also applicable under sampling setting, we also provide the sampling results in Appendix~\ref{sec:appvadiation}.
For all baselines, we apply at the same layer, search over the same number of parameters, and report the best-performing configuration for fair comparison.

\textbf{Implementation Details.}
We constrain \method and baseline methods on the residual stream following~\citep{hjer2025improving,galichin2025icoveredbaseshere} for fair comparison. 
More details are in Appendix~\ref{Sec:appimplementation}.

\textit{IPG settings.} 
For the policy gradient algorithm, we employ GRPO~\citep{shao2024deepseekmath}. The external verifier for reasoning outcomes comes from either rule function based (Appendix~\ref{Sec:appalgvariants}), such as generation length (Appendix~\ref{Sec:appresoninglength}) or reward model based signals (using Skywork-Reward-V2-Llama-3.1-8B \citep{liu2025skyworkrewardv2scalingpreferencedata}). 
For the baseline $\mathbf{h}'$ mentioned in~\cref{eq:ipg}, we set $\mathbf{h}'=\mathbf{0}\in\mathbb{R}^n$, following~\citep{dai-etal-2022-knowledge}. 
We also provide  analysis on    choices of baseline in Appendix~\ref{Sec:appsensibaseline}.
\vspace{-5pt}
\subsection{Causal Intervention on Localized Components (RQ1)}
\label{sec:reasoning_identicontrol}
We first validate whether the components localized by \method genuinely drive reasoning behaviors, rather than only correlate with successful outputs.

\label{sec:quantitative_control}

\begin{wraptable}{r}{0.52\linewidth}
\vspace{-15pt}
\centering
\small
\caption{Ablation of IPG-identified components.}
\resizebox{1.0\linewidth}{!}{%
\begin{tabular}{lccc}
\toprule
\textbf{Task} & \textbf{Original} & \textbf{Random} &\textbf{IPG Feature} \\
\midrule
Reasoning   & 63.00 &  59.80 {\scriptsize (-3.20)}  &28.80 {\scriptsize (-34.2)} \\
Non-Reasoning  & 12.11 & 9.81{\scriptsize (-2.30)}  & 10.81 {\scriptsize (-1.3)}\\
\bottomrule
\end{tabular}
}
\label{tab:ablation_neuron_feature}
\vspace{-12pt}
\end{wraptable}

\textbf{Finding 1: \method identifies components with profound sequential influence.} 
We examine whether \method-identified components serve as a \textit{outcome-aware mediator} in reasoning by (i) ablating them (setting their activation to 0) to test whether reasoning degrades, and (ii) amplifying them to test whether reasoning improves.
We ablate the top-ranked \method-identified features in \cref{tab:ablation_neuron_feature}. 
It shows a sharp drop on reasoning task (MATH-500~\citep{hendrycks2021measuringmathematicalproblemsolving}), with minor change on non-reasoning task (AlpacaEval~\citep{dubois2024lengthcontrolled}), indicating that the localized components are genuinely critical for reasoning.

We further evaluate \method alongside baseline methods across multiple reasoning datasets, applying  enhancement interventions. 
In \cref{tab:maintable_enhancing}, \method achieves consistent control over reasoning performance. 
Enhancing \method-identified components yields consistent gains, while integrating \method with Sparse Autoencoders (\method-SAE) pushes effectiveness even further. 
Baselines relying on text patterns (R.N., SAE-R) or contrastive pairs (C.V.) yield marginal or unstable improvements, suggesting they only capture short term effect rather than sequential influence required by multi-step reasoning. 

We additionally compare \method with three representative component-attribution methods, including Integrated Gradients~\citep{sundararajan2017axiomatic}, Neuron Shapley~\citep{neushapnips2020}, and AttnLRP~\citep{attenlrpicml2024}, in the neuron intervention setting on Qwen2.5-Math-1.5B-Instruct.
All three baselines improve accuracy over the original model, while \method achieves the highest accuracy across the four evaluated benchmarks.
These results extend the comparison beyond reasoning-specific baselines and support the utility of outcome-based sequential attribution.
The full results and attribution settings are provided in Appendix~\ref{app:attribution_baselines}.

\begin{table*}[t]
\small
\centering
\vspace{-10pt}
\caption{
Accuracy (\%) after enhancing components selected by \method and baseline methods across reasoning benchmarks under greedy decoding.
Results are reported for Qwen2.5-Math-1.5B-Instruct and Llama3.1-8B-Instruct.
Values following $\pm$ denote the half-width of the 95\% confidence interval, in percentage points.
The best and second-best point estimates in each column are shown in \textbf{bold} and \underline{underlined}, respectively; ties receive the same formatting.
}
\vspace{-5pt}
\resizebox{0.98\linewidth}{!}{%
\begin{tabular}{lcccccccc}
\toprule
\textbf{Method} &
\multicolumn{4}{c}{\textbf{Qwen2.5-Math-1.5B}} &
\multicolumn{4}{c}{\textbf{Llama3.1-8B}} \\
\cmidrule(lr){2-5} \cmidrule(lr){6-9}
& GSM8K & MATH & AIME & GPQA
& GSM8K & MATH & AIME & GPQA \\
\midrule
Original
& 82.41{\scriptsize$\pm$0.65}
& 63.00{\scriptsize$\pm$0.95}
& 10.00{\scriptsize$\pm$3.17}
& 24.75{\scriptsize$\pm$1.92}
& 85.89{\scriptsize$\pm$0.72}
& 41.80{\scriptsize$\pm$0.29}
& 6.67{\scriptsize$\pm$3.17}
& 23.23{\scriptsize$\pm$0.96} \\

Random
& 82.86{\scriptsize$\pm$0.65}
& 63.20{\scriptsize$\pm$0.88}
& \underline{16.67}{\scriptsize$\pm$4.75}
& 26.77{\scriptsize$\pm$1.68}
& 86.20{\scriptsize$\pm$0.65}
& 41.40{\scriptsize$\pm$0.29}
& 6.67{\scriptsize$\pm$4.75}
& 27.78{\scriptsize$\pm$0.96} \\

Activation
& 82.56{\scriptsize$\pm$0.65}
& 63.40{\scriptsize$\pm$1.33}
& \textbf{20.00}{\scriptsize$\pm$4.75}
& 25.75{\scriptsize$\pm$1.92}
& 85.75{\scriptsize$\pm$0.61}
& \underline{43.00}{\scriptsize$\pm$0.19}
& \underline{13.33}{\scriptsize$\pm$4.75}
& 25.25{\scriptsize$\pm$0.72} \\

R.N.~\citep{rai-yao-2024-investigation}
& 82.26{\scriptsize$\pm$0.83}
& 62.20{\scriptsize$\pm$1.14}
& 13.33{\scriptsize$\pm$4.62}
& 24.24{\scriptsize$\pm$1.92}
& 86.05{\scriptsize$\pm$0.65}
& 40.80{\scriptsize$\pm$1.23}
& 6.67{\scriptsize$\pm$4.75}
& 21.21{\scriptsize$\pm$2.16} \\

C.V.~\citep{hjer2025improving}
& 83.85{\scriptsize$\pm$0.07}
& 62.60{\scriptsize$\pm$1.04}
& \underline{16.67}{\scriptsize$\pm$3.17}
& 28.78{\scriptsize$\pm$1.44}
& 85.97{\scriptsize$\pm$0.86}
& 40.00{\scriptsize$\pm$0.57}
& 6.67{\scriptsize$\pm$4.75}
& \underline{28.78}{\scriptsize$\pm$0.48} \\

SAE-R~\citep{galichin2025icoveredbaseshere}
& 83.32{\scriptsize$\pm$0.40}
& 64.00{\scriptsize$\pm$0.76}
& 13.33{\scriptsize$\pm$4.75}
& 25.76{\scriptsize$\pm$1.68}
& 86.28{\scriptsize$\pm$0.54}
& 41.80{\scriptsize$\pm$0.85}
& 10.00{\scriptsize$\pm$3.17}
& 26.77{\scriptsize$\pm$0.96} \\

\textbf{\method (Ours)}
& \underline{84.61}{\scriptsize$\pm$0.40}
& \underline{65.80}{\scriptsize$\pm$0.88}
& \textbf{20.00}{\scriptsize$\pm$3.17}
& \underline{33.33}{\scriptsize$\pm$1.68}
& \underline{87.18}{\scriptsize$\pm$0.43}
& 42.40{\scriptsize$\pm$0.66}
& \underline{13.33}{\scriptsize$\pm$4.62}
& 27.27{\scriptsize$\pm$0.96} \\

\textbf{\method-SAE (Ours)}
& \textbf{84.76}{\scriptsize$\pm$0.50}
& \textbf{66.00}{\scriptsize$\pm$0.76}
& \textbf{20.00}{\scriptsize$\pm$4.75}
& \textbf{35.35}{\scriptsize$\pm$1.20}
& \textbf{87.41}{\scriptsize$\pm$0.43}
& \textbf{44.00}{\scriptsize$\pm$0.57}
& \textbf{20.00}{\scriptsize$\pm$4.62}
& \textbf{29.80}{\scriptsize$\pm$0.72} \\
\bottomrule
\end{tabular}%
}
\label{tab:maintable_enhancing}
\vspace{-20pt}
\end{table*}

\textbf{Finding 2: \method alignments with a direct causal mediation proxy.}
To complement the enhancement experiments, we measured the alignment (overlap ratio) between the top neurons identified by our method with a direct mediation proxy (details in Appendix~\ref{Sec:appevalmetric}).
Let $T_{\mathrm{CMA}}^{k}$ denote the reference set of $k$ neuron indices
obtained from this proxy, and let $T_m^{k}$ denote the top-$k$ neuron
indices selected by attribution method $m$.
We measure their alignment as $\operatorname{Overlap@}k(m)
=
\frac{
\left|T_{\mathrm{CMA}}^{k} \cap T_m^{k}\right|
}{k}$.

As shown in Table~\ref{tab:mediation_proxy}, \method achieves a
substantially higher mean overlap than the attribution baselines
on both models.
These results suggest that \method provides an approximation of causal mediation.
\begin{wraptable}{r}{0.58\linewidth}
\centering
\vspace{-0pt}
\caption{
Alignment with a group-based zero-ablation mediation proxy on
a subset of MATH-500.
Values are mean $\operatorname{Overlap@}k \pm$ standard deviation with $k$ corresponding to approximately
5\% of neurons.
Qwen and DeepSeek denote Qwen2.5-Math-1.5B-Instruct and
DeepSeek-R1-Distill-Qwen-1.5B, respectively.
}
\label{tab:mediation_proxy}
\resizebox{1.0\linewidth}{!}{%
\begin{tabular}{lcc}
\toprule
\textbf{Method} & \textbf{Qwen} & \textbf{DeepSeek} \\
\midrule
Random         & $0.064 \pm 0.023$ & $0.064 \pm 0.023$ \\
IG~\citep{sundararajan2017axiomatic}             & $0.091 \pm 0.025$ & $0.066 \pm 0.022$ \\
Neuron SHAP~\citep{neushapnips2020} & $0.072 \pm 0.020$ & $0.065 \pm 0.005$ \\
AttnLRP~\citep{attenlrpicml2024}        & $0.082 \pm 0.018$ & $0.062 \pm 0.013$ \\
\textbf{\method (Ours)}
& $\mathbf{0.305} \pm 0.113$
& $\mathbf{0.246} \pm 0.030$ \\
\bottomrule
\end{tabular}
}
\vspace{-10pt}
\end{wraptable}

\begin{figure*}[htp]
    \vspace{-12pt}
    \centering
    \includegraphics[width=0.9\linewidth]{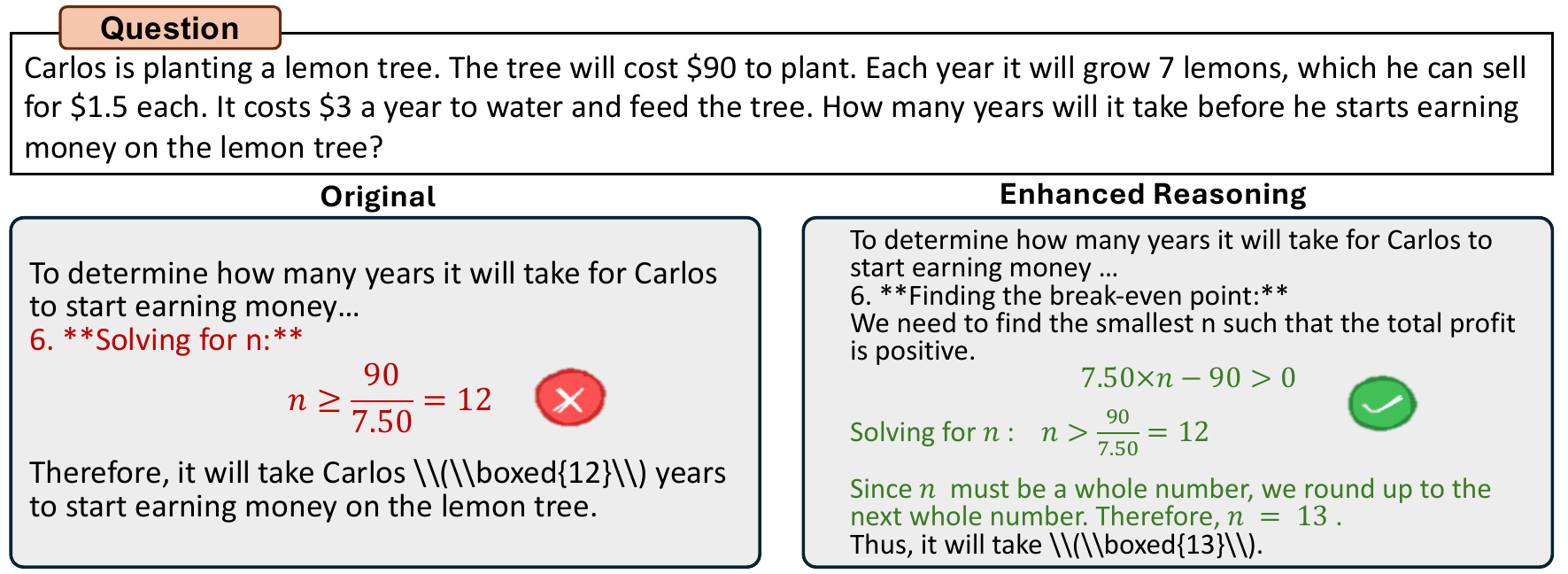}
    \caption{Responses generated by Qwen2.5-Math-1.5B-Instruct~\citep{yang2024qwen25mathtechnicalreportmathematical} on one example in GSM8K~\citep{cobbe2021trainingverifierssolvemath}, 
    including both original and intervened outputs, with the model's reasoning ability elicited to arrive at the correct answer. 
    More examples are in Appendix~\ref{Sec:appcasegenerateresponses}.
    }
    \vspace{-20pt}
    \label{fig:caseingsm8k}
\end{figure*}

\textbf{Finding 3: \method refines the entire reasoning trajectory.}
To understand how interventions alter the generation process, we qualitatively compare responses from Qwen2.5-Math-1.5B-Instruct~\citep{yang2024qwen25mathtechnicalreportmathematical} on GSM8K before and after intervening on \method-identified components. 
As illustrated in \cref{fig:caseingsm8k}, the baseline model fails to follow strict logical constraints. 
Amplifying \method components does not merely force the correct final answer; instead, it fundamentally corrects the intermediate reasoning steps, resulting in a more coherent and rigorous logical trajectory.

\begin{figure}[htbp]
    \centering
    \begin{subfigure}{0.49\columnwidth}
        \centering
        \includegraphics[width=1.0\columnwidth]{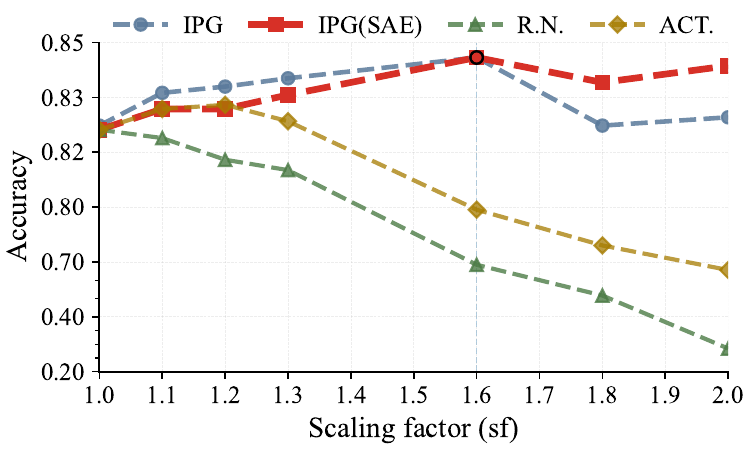} 
        \caption{Scaling factor effects.
        }
        \label{fig:sf}
    \end{subfigure}
    \hfill
    \begin{subfigure}{0.49\columnwidth}
        \centering
        \includegraphics[width=1.0\columnwidth]{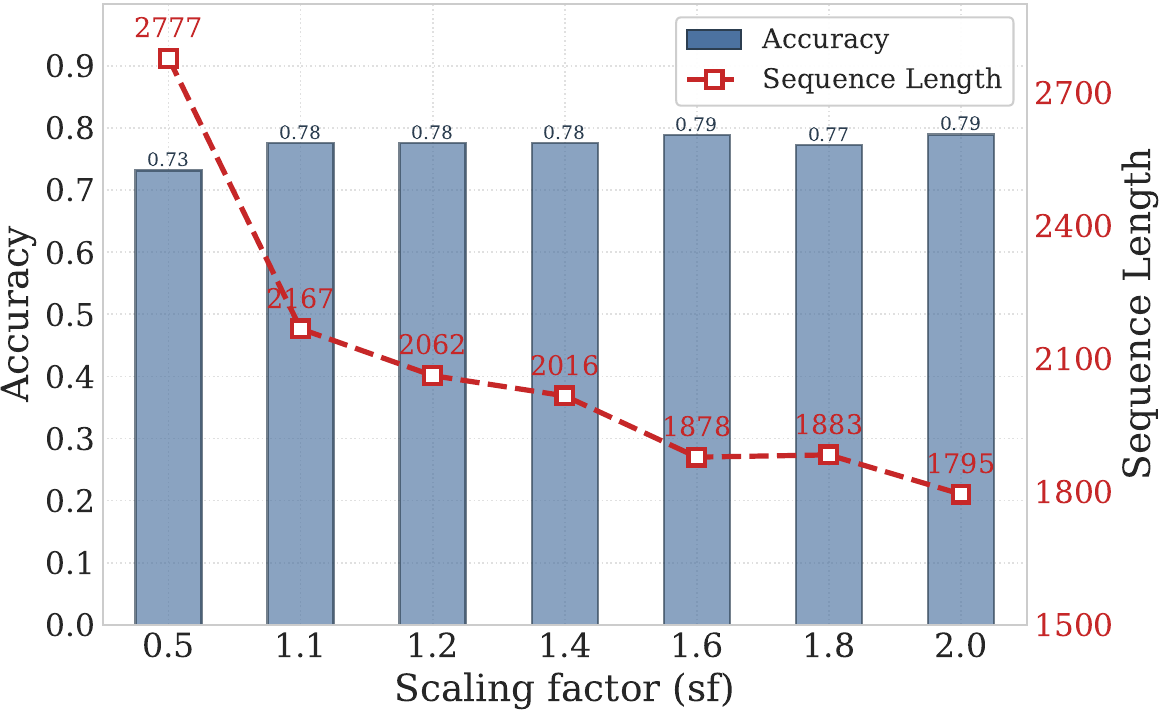}
        \caption{Reasoning strength effects.}
        \label{fig:lengthcontrol}
    \end{subfigure}
    \vspace{-5pt}
    \caption{
    Effects of targeted interventions on reasoning performance.
    Left: impact of the intervention scaling factor on Qwen2.5-Math-1.5B-Instruct~\citep{yang2024qwen25mathtechnicalreportmathematical}.
    Right: Reasoning strength control in GSM8K~\citep{cobbe2021trainingverifierssolvemath} for DeepSeek-R1-Distilled-Qwen-1.5B~\citep{guo2025deepseek}.
    }
    \label{fig:scaling_effects}
    \vspace{-5pt}
\end{figure}

\textbf{Finding 4: \method enables predictable and granular modulation.} 
We evaluate the precise controllability of two facets of reasoning: \textit{capability} (performance sensitivity) and \textit{strength} (reasoning chain length). By varying the intervention scaling factor $\gamma$, \method yields smooth, predictable control over reasoning capability, whereas baseline methods exhibit unstable dynamics (\cref{fig:sf}). 
Furthermore, for reasoning strength, \method successfully modulates the length of the reasoning trajectory without sacrificing task accuracy (\cref{fig:lengthcontrol}), suggesting that \method isolates the distinct internal components governing different dimensions of reasoning behavior.

\subsection{Generalizability of Reasoning Mechanisms (RQ2)}
\label{sec:generalizability}

A key criterion for mechanistic discovery is \textit{generalization}: components that genuinely implement reasoning should not be tied to a single dataset and domain.
We therefore examine whether \method-identified components transfer beyond the source task on which they are localized.
Specifically, we evaluate both \emph{cross-dataset transfer} within mathematical tasks and \emph{cross-domain transfer} across mathematical, coding, and commonsense reasoning tasks.

\textbf{Finding 5: \method identifies reasoning mechanisms with cross-dataset and cross-domain consistency.} 
Components localized on MATH500~\citep{hendrycks2021measuringmathematicalproblemsolving} transfer (directly intervene) effectively to GSM8K~\citep{cobbe2021trainingverifierssolvemath}, outperforming baselines in cross-dataset generalization. Remarkably, this efficacy extends across distinct domains, including coding (HumanEval) and commonsense reasoning (CSQA) (\cref{tab:crossdomain}). 
Supported by Dice similarity metrics (\cref{fig:jaccard}, Appendix~\ref{Sec:appgeneralmechan}), these results suggest that \method captures core and transferable reasoning 
ability.

\begin{table*}[tbp]
\vspace{-5pt}
\small 
\centering
\caption{Cross-dataset (mathematics) and cross-domain (code and commonsense) transfer under greedy decoding. Best and second best results are shown in \textbf{bold} and \underline{underlined} format. We provide the half-width 95\% CI for our method.}
\vspace{-5pt}
\resizebox{\linewidth}{!}{ 
\begin{tabular}{l c c c c c c}
\toprule
 & \multicolumn{2}{c}{\textbf{Cross-Dataset}} & \multicolumn{4}{c}{\textbf{Cross-Domain}} \\
\cmidrule(lr){2-3} \cmidrule(lr){4-7}
 & \multicolumn{2}{c}{\textbf{MATH-500 $\to$ GSM8K}} & \multicolumn{2}{c}{\textbf{MATH-500 $\to$ HumanEval}} & \multicolumn{2}{c}{\textbf{HumanEval $\to$ CSQA}} \\
\cmidrule(lr){2-3} \cmidrule(lr){4-5} \cmidrule(lr){6-7}
\textbf{Method} & \textbf{Acc. (\%)}~$\uparrow$ & \textbf{Avg. Tok.} & \textbf{Acc. (\%)}~$\uparrow$ & \textbf{Avg. Tok.} & \textbf{Acc. (\%)}~$\uparrow$ & \textbf{Avg. Tok.} \\
\midrule
Original & 82.41 & 319.68 & 11.58 & 212.12 & 21.14 & 317.05 \\
Activation & 82.94 & 317.17 & 12.80 & 161.93 & 20.00 & 276.58 \\
R.N.~\citep{rai-yao-2024-investigation} & 82.87 & 312.63 & \underline{15.24} & 163.98 & 20.26 & 258.78 \\
C.V.~\citep{hjer2025improving} & \underline{83.39} & 320.03 & 14.02 & 195.33 & \underline{20.96} & 275.47 \\
\textbf{\method (Ours)} 
& \textbf{84.08}{\scriptsize$\pm$0.32}  
& 315.86 
& \textbf{21.34}{\scriptsize$\pm$1.45} 
& 191.40 
& \textbf{22.72}{\scriptsize$\pm$0.54}  
& 325.89 \\
\bottomrule
\end{tabular}
}
\vspace{-10pt}
\label{tab:crossdomain}
\end{table*}

\subsection{Deeper Mechanistic Insights (RQ3)}
\label{sec:mechanistic_insights}
\begin{figure}[bp]
    \centering
    \vspace{-15pt}
    \begin{subfigure}[b]{0.58\textwidth}
        \centering
        \includegraphics[width=\linewidth]{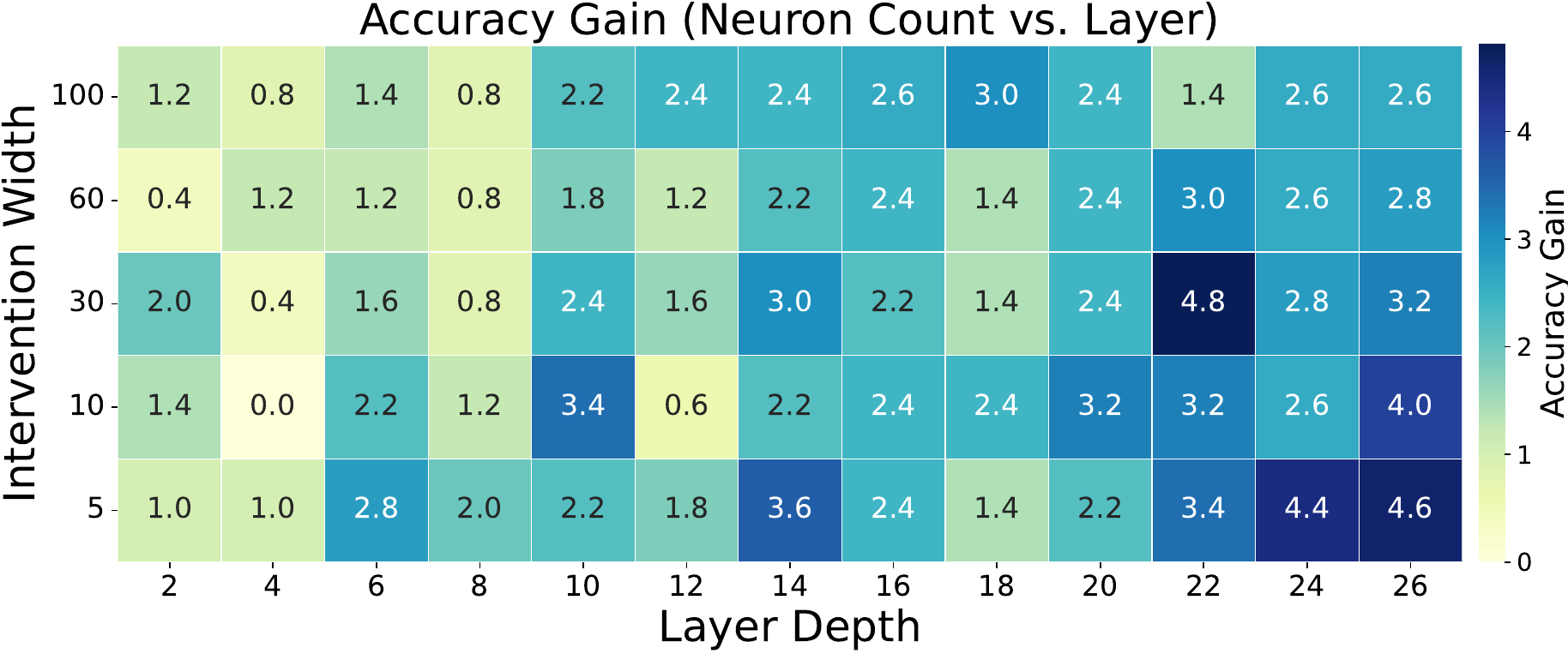}
        \caption{Performance Gain by Depth and Width}
        \label{fig:layer_heatmap}
    \end{subfigure}\hfill
    \begin{subfigure}[b]{0.38\textwidth}
        \centering

        \includegraphics[width=\linewidth]{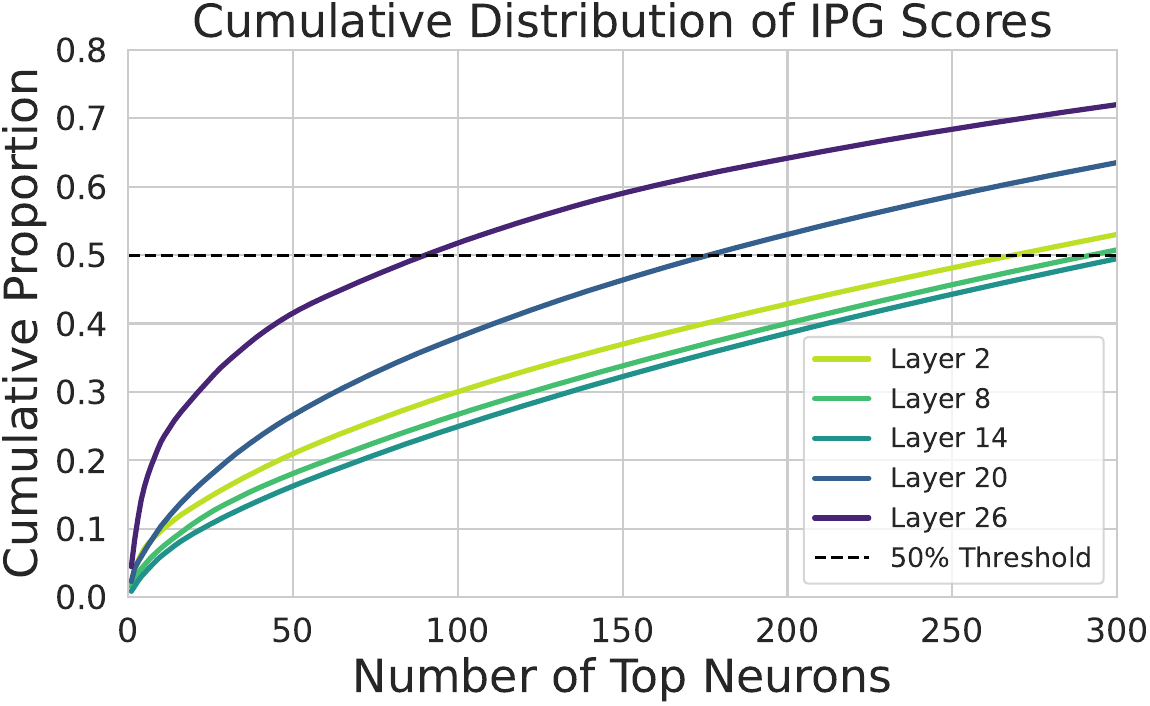}
        \caption{\method Score Cumulation}
        \label{fig:layer_cdf}
    \end{subfigure}
    
    \vspace{5pt}
    \caption{Layer-wise localization and concentration of reasoning components. 
    \textbf{(a)} The heatmap for reasoning task (MATH-500) performance enhancement across different network layers and intervention widths (number of targeted neurons). 
    \textbf{(b)} The cumulative \method\ scores for selected layers.}
    \label{fig:layer_analysis}
    \vspace{-10pt}
\end{figure}

Having established the efficacy and generalizability of the identified components, we leverage \method to unpack the mechanisms of LLM reasoning at a granular level. Specifically, we investigate where these components are located (~\cref{subsec: where}), what reasoning functions they encode (~\cref{subsec: what}), and when they contribute to the reasoning process (~\cref{subsec: when}).

\subsubsection{Spatial Localization and Concentration}
\label{subsec: where}
\textbf{Finding 6: Reasoning-related components are spatially localized and concentrated.}
We first examine \emph{where} influential components are located.
As shown in \cref{fig:layer_heatmap}, performance gains vary across
layers and intervention widths, with the largest gains concentrated
in specific middle-to-late layers.
This pattern suggests that components with substantial effects on
reasoning performance are not uniformly distributed throughout
the model.

The cumulative attribution scores in \cref{fig:layer_cdf}, computed
by summing $S_i$ over selected components $\mathbf{P}$
in \cref{eq:ipgscore}, further show that a relatively small subset
of components accounts for a large share of attribution importance
within these layers.
Together, the attribution concentration and intervention gains
identify localized regions whose components are particularly
influential for reasoning.

\subsubsection{Semantic Analysis}
\label{subsec: what}

\textbf{Finding 7: \method-selected features exhibit distinct
reasoning-related semantic patterns.}
Having identified where influential components are concentrated,
we next examine \emph{what} they represent.
Individual neurons can entangle multiple concepts
\citep{mu2020,Olah2020ZoomIn}, so we analyze \method-selected
features in the Sparse Autoencoder (SAE) representation space.
For each feature, we collect its highest-activation contexts
across GSM8K.

\begin{figure}[t]
    \centering
    \begin{minipage}[b]{0.48\textwidth}
        \centering
        \small
        \label{tab:token_analysis}
        \begin{tabular}{c p{0.65\linewidth}}
        \toprule
        \textbf{Feat. ID} & \textbf{Peak Context} \\
        \midrule
        486 & Let's start with the \textbf{first step}:\\ \addlinespace
        9889 & Therefore, \textbf{the answer} is \(\boxed{14}\).\\ \addlinespace
        7349 & 25\% of 80 is \textbf{0.25} * \textbf{80} = 20,\\ \addlinespace
        12179 & \textbf{Therefore}, the total amount of money Maddy will spend... \\ \addlinespace
        11155 & The \textbf{total number} of rows of flowers Jackson will plant is... \\
        \bottomrule
        \end{tabular}
        \captionof{table}{SAE features identified by \method alongside the text contexts. Tokens receiving the maximal activation value are in \textbf{bold}.}
        \label{tab:tokenact}
    \end{minipage}
    \hfill
    \begin{minipage}[b]{0.48\textwidth}
        \centering
        \includegraphics[width=\linewidth]{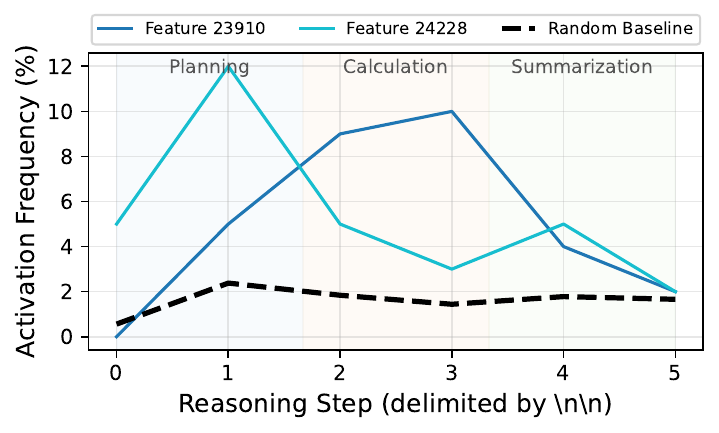}
        \captionof{figure}{Temporal activation dynamics of SAE features, measured across normalized reasoning progress discretized into step deciles.}
        \label{fig:dynamics}
    \end{minipage}
    
    \vspace{-5pt}
\end{figure}

As shown in \cref{tab:tokenact}, these contexts reveal distinct
reasoning-related patterns.
Feature~\#7349 activates during explicit arithmetic calculations,
whereas Features~\#486 and \#12179 appear in planning and
summarization contexts, respectively.
Additional examples are provided in
Appendix~\ref{Sec:appcasecompdisentangle}.
These differences suggest that the selected features are
associated with complementary aspects of problem solving, providing interpretable descriptions
of the selected components.
\subsubsection{Temporal Organization}
\label{subsec: when}

\textbf{Finding 8: \method-selected features exhibit distinct
temporal activation patterns.}
Finally, we examine \emph{when} these features are active.
We track representative SAE features across reasoning trajectories,
grouping their activations into step deciles of normalized generation
progress.
As shown in \cref{fig:dynamics}, different features exhibit
distinct activation profiles, with activity concentrated at
different stages rather than uniformly distributed throughout
generation.

These temporal differences complement the semantic patterns
in \cref{tab:tokenact}: the selected features differ not only
in their activating contexts, but also in when they are engaged
during a generated solution (e.g., planning or calculating).
Overall, the spatial, semantic, and temporal analyses characterize
where influential components are concentrated, what patterns
they are associated with, and when they become active.

\subsubsection{Mechanisms Under Distillation}
\label{sec:deepseek-distilled}

\textbf{Finding 9: Distillation preserves and amplifies base mechanisms.} 
Finally, we transfer \method-identified neurons from a standard prompt-elicited model (Qwen2.5-Math-1.5B-Instruct) directly to its reasoning-distilled counterpart (DeepSeek-R1-Distilled-Qwen-1.5B), \textbf{without} re-running \method.
As shown in \cref{tab:distill_steer}, enhancing the inherited neurons within the DeepSeek-distilled model consistently improves reasoning performance. 
This insight suggests that modern model distillation may preserve the internal architecture for reasoning pathways and it relies on and strengthens the fundamental reasoning mechanisms in the base model that are already localized by \method.
\begin{table}
\centering
\vspace{-0pt}
\caption{
Enhancing neurons from prompt-elicited to reason-distilled models without re-running \method.
}
\resizebox{0.9\linewidth}{!}{
\begin{tabular}{lcccc}
\toprule
\textbf{Model} & \textbf{GSM8K} & \textbf{MATH500} & \textbf{AIME2024} & \textbf{AIME2025} \\
\midrule
DeepSeek-R1-Distill-Qwen-1.5B & 75.21 & 60.60 & 16.67 & 13.33\\
+ IPG Neuron Transferring    & \textbf{78.01} & \textbf{63.80} & \textbf{30.00} &\textbf{30.00}\\
\bottomrule
\end{tabular}
}
\label{tab:distill_steer}
\vspace{-5pt}
\end{table}
\section{Conclusion}
In this work, we introduced Integrated Policy Gradient (\method), a novel, training-free framework for interpreting LLM reasoning guided by \textit{outcome-oriented} and \textit{sequential-influence-aware} principles. 
By bridging Causal Mediation Analysis with policy gradients~\citep{schulman2017proximal} and path integration, \method successfully back-propagates sparse, non-differentiable outcome signals (e.g., final task correctness) to isolate outcome-relevant mediators across long-horizon trajectories. 
Empirically, our evaluations revealed that complex reasoning relies on concentrated regions and exhibits specialized behaviors. 
Beyond enabling predictable and granular control over reasoning behavior, \method demonstrated robust cross-dataset and cross-domain transferability. 
Future work can explore extending this framework to other domains such as emotional intelligence and creativity and further refining the manipulation of these localized components.

\begin{ack}
The research was supported by the National Natural Science Foundation of China (Grant No. 62406193) and the ShanghaiTech AI Initiative (Grant No. AI2026B08).
The authors gratefully acknowledge further assistance from the Key Laboratory of Intelligent Perception and Human-Machine Collaboration (Shanghaitech University), Ministry of Education, and the HPC Platform of ShanghaiTech University.
This research was also supported by Meituan.

There are no competing interests to declare.
\end{ack}

\bibliographystyle{plainnat}
\bibliography{main}

\newpage
\appendix
\section{Algorithms of \method}
\label{Sec:appalgoipg}
In this section, we provide two main algorithms of our method in order to clearly illustrate the key steps of \method. The first algorithm focuses on \emph{identifying reasoning-critical components} by attributing outcome-weighted signals to internal representations. The second algorithm is based on these identified components to \emph{control reasoning behavior} of the model.  (see more details in~\cref{Sec:ipgattribution}).

\subsection{Identifying Reasoning Components}
We introduce how to identify reasoning-critical components. We use a support set of $M$ samples, which can be either a subset or the full dataset of any reasoning-related benchmarks like GSM8K \citep{cobbe2021trainingverifierssolvemath}. We denote output residual stream of transformer block as $\mathcal{F}(\cdot)$ which takes a sample as input and we can get the hidden state of residual stream. 
Here, $\Phi$ can be either the identity map, corresponding to neuron-level IPG, or the SAE encoder, corresponding to feature-level IPG. After that, we can localize the reasoning-critical components.
For the calculation of attribution, we use Riemann approximation to estimate integral:

\begin{equation}
\label{Eq:attr_discrete_correct}
\mathrm{IPG}(i;\bx)
=
\sum_{t=1}^{T} (h_{t,i}-h'_{t,i})
\cdot
\frac{1}{q}\sum_{k=1}^{q}
\mathbb{E}_{\tau\sim\pi_{\theta}}
\!\left[
\frac{\partial}{\partial h_{t,i}}
\log \pi_{\theta}\!\left(
a_t \mid s_t;\,
\mathbf{h}'_{t} + \frac{k}{q}\bigl(\mathbf{h}_{t}-\mathbf{h}'_{t}\bigr)
\right)
\cdot
A^{\pi}\!\left(s_t,a_t\right)
\right].
\end{equation}

where $\mathbf{h}'=[h_1',\ldots h'_m]$ is a relative baseline of $\mathbf{h}$ and $q$ is the number of discrete steps or partitions used in the Riemann approximation to estimate the integral. $\bx$ is the data sample. 
Notably, we use $\mathbf{h}$ in the following~\cref{alg:identify} and \cref{alg:control} to denote the representation of either neuron or SAE feature. 
More details about advantage $A^\pi$ are provided in~\cref{Sec:appalgvariants}.

\begin{algorithm}[htbp]
\caption{\method: Identifying Reasoning Components}
\label{alg:identify}
\begin{algorithmic}[1]
\REQUIRE Support set $X = \{\mathbf{x}_1, \ldots, \mathbf{x}_M\}$, number of components $p \in \mathbb{N}$
\ENSURE Selected component indices $\mathbf{P}$

\FOR{$j = 1$ to $M$}
    \STATE Extract hidden states $\mathbf{z} = [z_1, z_2, \ldots, z_n]$ from $\mathcal{F}(\mathbf{x}_j)$
    \STATE Compute components $\mathbf{h} = [h_1, h_2, \ldots, h_d] = \Phi(\mathbf{z})$
    \FOR{$i = 1$ to $m$}
        \STATE $S(h_i; X) \leftarrow S(h_i; X) + \mathrm{IPG}(i; \mathbf{x}_j)$ \COMMENT{\cref{eq:ipg}}
    \ENDFOR
\ENDFOR

\STATE $\mathbf{P} \leftarrow \operatorname*{arg\,top\text{-}p}_{i \in [m]} \; S(u_i; X)$ \COMMENT{\cref{eq:ipgscore}}
\STATE \textbf{return} $\mathbf{P}$
\end{algorithmic}
\end{algorithm}

\begin{algorithm}[htbp]
\caption{IPG: Controlling Reasoning Behavior over Generation Steps}
\label{alg:control}
\begin{algorithmic}[1]
\REQUIRE 
Component indices $P = \{i_1, \ldots, i_p\}$, 
hidden states $\{\mathbf{z}_t\}_{t=1}^{T}$, 
scaling factor $\gamma \in \mathbb{R}$
\ENSURE Modified hidden states $\{\mathbf{z}_t\}_{t=1}^{T}$

\FOR{$t = 1$ to $T$} 
    \STATE Compute components 
    $\mathbf{h}_t = [h_{t,1}, \ldots, h_{t,d}] = \Phi(\mathbf{z}_t)$ \COMMENT{\cref{Eq:encode}}
    \STATE Initialize $\mathbf{h}'_t \leftarrow \mathbf{h}_t$
    \FOR{$k = 1$ to $p$} 
        \STATE $h'_{t,i_k} \leftarrow \gamma \cdot h'_{t,i_k}$ \COMMENT{Scaling,~\cref{Eq:intervene}}
    \ENDFOR
    
    \IF{$\Phi$ is Identity}
        \STATE $\mathbf{z}_t \leftarrow \mathbf{h}'_t$
    \ELSIF{$\Phi$ is SAE-encoder}
        \STATE $\boldsymbol{\epsilon}_t \leftarrow 
        \mathbf{z}_t - (\mathbf{W}_{\mathrm{dec}}\mathbf{h}_t + \mathbf{b}_{\mathrm{dec}})$
        \STATE $\mathbf{z}_t \leftarrow 
        \mathbf{W}_{\mathrm{dec}}\mathbf{h}'_t + \mathbf{b}_{\mathrm{dec}} + \boldsymbol{\epsilon}_t$
        \COMMENT{\cref{Eq:encode}}
    \ENDIF
\ENDFOR

\STATE \textbf{Return} $\{\mathbf{z}_t\}_{t=1}^{T}$
\end{algorithmic}
\end{algorithm}








\subsection{Controlling Reasoning Behavior}
After identifying reasoning-critical components, we further demonstrate how to directly control the reasoning behavior of the model. 
Given a set of target component indexes $\mathbf{P}$, we first encode the hidden state $\mathbf{h}$ into the representation space $\mathbf{s}$ using $\Phi$, which can be either the identity mapping (for neuron-level control) or the SAE encoder (for feature-level control). 
We then intervene on the selected components by scaling their activations with a factor $\gamma$, yielding a modified representation $\mathbf{h}'$. Finally, we decode $\mathbf{s}’$ back to the residual stream. 
Notably, in the SAE setting, the reconstruction also retains the error term to ensure faithful recovery of the original hidden state structure (see more details in~\cref{Sec:apptrainsae} about this).

\subsection{Variants of \method}
\label{Sec:appalgvariants}
In this section, we provide variants of our \method that support different algorithms of policy gradient~\citep{mnih2016asynchronous}, as mentioned in Section~\ref{Sec:ipgattribution}.

In GRPO~\citep{shao2024deepseekmath}, the structure of \method remains identical, except the advantage $A_t$ is replaced with a group-relative advantage:
\begin{equation}
A_{t} = \frac{r_{t} - \text{mean}(\{r_1, r_2, \dots, r_N\})}{\text{std}(\{r_1, r_2, \dots, r_N\})}.
\end{equation}
where $T$ is the sequence length, $N$ is the sample number, and $\pi_\theta$ is the policy parameterized by $\theta$, i.e., the LLM to be interpreted.
We treat the value of signal function $J(\cdot)$ as the reward $r_t$.
$s_t^{(k)}$ is the prefix of the sequence $k$ until $t$ steps, $a_t^{(k)}$ is the next token in the sequence $k$, and $A^\pi(s_t, a_t)$ is the advantage function that estimates benefit of selecting $a_t$ over the baseline, often derived from rewards like reasoning accuracy scores~\citep{zhong2025comprehensive}.
\begin{equation}
\small
\label{eq:grpoipg}
\mathrm{IPG}^{\mathrm{GRPO}}(i;\bx)
=
\sum_{t=1}^{T} (h_{t,i}-h'_{t,i})
\int_{0}^{1}
\frac{1}{N}\sum_{n=1}^{N}
\left[
\frac{\partial}{\partial h_{t,i}}
\log \pi_\theta\!\left(
a_t^{(n)} \mid s_t^{(n)};\,
\mathbf h'_t + \alpha(\mathbf h_t-\mathbf h'_t)
\right)
\cdot
A_{t}^{\mathrm{GRPO},(n)}
\right]
\, d\alpha .
\end{equation}

Here, $A_t$ denotes the advantage function, which measures how much better the chosen action $a_t$ is compared to the expected baseline at state $s_t$. 
$\bx$ is the data sample.
As we can see in~\cref{eq:grpoipg}, our \method can be seamlessly integrated with diverse policy gradient algorithms by integrating their advantage calculation.

For the signal function $J(\cdot)$, we support both \textbf{rule-based} and \textbf{model-based} signals.  
In the rule-based setting, the reward $r$ is defined as a binary indicator of final-answer correctness:
\begin{equation}
    J = \mathbb{I}\{\hat{y} = y\} \;=\; 
\begin{cases}
1, & \text{if}\; \hat{y} = y, \\
0, & \text{otherwise}.
\end{cases}
\end{equation}
In contrast, model-based rewards provide a continuous evaluation of reasoning quality.  
For example, given a reward model $M(\cdot)$, we can assign rewards as
\begin{equation}
    J = M(x, \hat{y}),
\end{equation}
where $M$ may return partial credit (e.g., proportion of correct intermediate steps) or preference scores from human-aligned models.  
This yields a richer supervision signal than binary correctness, enabling finer-grained attribution of reasoning ability.
A comparison of these reward sources is presented in Appendix~\ref{Sec:apprewardsig}.

\subsection{Formal Justification of IPG}
\label{Sec:appjustification}
\subsubsection{Extending Policy Gradient to Representation Space}
\label{Sec:appjustificationpg}
In this section, we provide a rigorous justification for~\cref{eq:policygrad}, which computes the policy gradient with respect to a deterministic hidden activation $\mathbf{h}$ rather than the policy parameters.

We consider an episodic Markov decision process and define the expected return as
\begin{equation}
J(\mathbf{h}) = \mathbb{E}_{\tau \sim \pi_\theta(\cdot \, ; \mathbf{h_t})} [ R(\tau) ]
= \int \pi_\theta(\tau \, ; \mathbf{h_t}) R(\tau) \, d\tau ,
\end{equation}
where $\tau = (s_1, a_1, \ldots, s_T, a_T)$ denotes a trajectory (including $s_t$ for cleaner proof), $R(\tau)$ is the total return, and $\mathbf{h_t}$ is a deterministic hidden activation that influences the policy $\pi_\theta$.

Assuming standard regularity conditions, we differentiate $J(\mathbf{h})$ with respect to $\mathbf{h}$ under the integral:
\begin{equation}
\frac{\partial J}{\partial \mathbf{h}}
= \int \frac{\partial \pi_\theta(\tau \, ; \mathbf{h_t})}{\partial \mathbf{h}} R(\tau) \, d\tau .
\end{equation}

Applying the score-function (log-derivative) identity,
\begin{equation}
\frac{\partial \pi_\theta(\tau \, ; \mathbf{h_t})}{\partial \mathbf{h}}
= \pi_\theta(\tau \, ; \mathbf{h_t}) \frac{\partial}{\partial \mathbf{h}} \log \pi_\theta(\tau \, ; \mathbf{h_t}),
\end{equation}
we obtain
\begin{equation}
\frac{\partial J}{\partial \mathbf{h}}
= \mathbb{E}_{\tau \sim \pi_\theta}
\left[
\frac{\partial}{\partial \mathbf{h}} \log \pi_\theta(\tau \, ; \mathbf{h_t}) \cdot R(\tau)
\right].
\end{equation}

Since the trajectory probability factorizes as
\begin{equation}
\log \pi_\theta(\tau \, ; \mathbf{h_t})
= \sum_{t=1}^{T} \log \pi_\theta(a_t \mid s_t ; \mathbf{h}_t),
\end{equation}
the gradient can be written as
\begin{equation}
\frac{\partial J}{\partial \mathbf{h}}
= \mathbb{E}_{\tau \sim \pi_\theta}
\left[
\sum_{t=1}^{T}
\frac{\partial}{\partial \mathbf{h_t}}
\log \pi_\theta(a_t \mid s_t ; \mathbf{h}_t)
\cdot R(\tau)
\right].
\end{equation}

For any baseline $b(s_t)$ independent of the action $a_t$, we have
\begin{equation}
\mathbb{E}_{\tau \sim \pi_\theta}
\left[
\frac{\partial}{\partial \mathbf{h}}
\log \pi_\theta(a_t \mid s_t ; \mathbf{h}_t) \cdot b(s_t)
\right] = 0,
\end{equation}
since $\int \pi_\theta(a_t \mid s_t ; \mathbf{h}_t)\, da_t = 1$ for all $\mathbf{h}_t$.
Subtracting this zero term yields the advantage-based form:
\begin{equation}
\small
\frac{\partial J}{\partial \mathbf{h}}
=\;
\mathbb{E}_{\tau\sim\pi_{\theta}}\!\left[\sum_{t=1}^{T}
\frac{\partial}{\partial \mathbf{h_t}}\log\pi_{\theta}\big(a_t\mid s_t;\mathbf{h}_t\big)
\cdot A^{\pi}(s_t,a_t)\right],
\end{equation}
which is~\cref{eq:policygrad}.

\paragraph{Discussion.}
This derivation shows that the policy gradient formulation naturally extends to deterministic hidden activations. The gradient measures how perturbations to internal representations influence the expected return through their effect on the policy, thereby justifying the use of policy-gradient-based attribution for internal components.

\subsubsection{Integrate Policy Gradient with Path Integral}
\label{app:ig_justification}

In this section, we derive the Integrated Policy Gradient (IPG). 
Let $\bx$ be the input prompt and $\tau$ be the generated trajectory with probability $\pi_\theta(\tau \mid \bx)$.
We define the attribution for an internal representation vector $h$ relative to a baseline $h'$ by considering the interpolated state $h(\alpha) := h' + \alpha(h - h')$ for $\alpha \in [0, 1]$~\citep{sundararajan2017axiomatic,dhamdhere2018how}.
The choices of baseline $h'$ are usually counterfactual values including $\mathbb{E}_{x\in\mathcal{D}}[h_i]$ or just zero value, serving as mean or zero ablation~\citep{cmasurveyma2026}.
We also provide our intervention results in Appendix~\ref{Sec:appsensibaseline}, which suggests our robustness under different choices of baseline value.

We define the objective function $\mathcal{J}(\alpha)$ as the \textit{expected reward} conditioned on this interpolated state: $\mathcal{J}(\alpha) := \mathbb{E}_{\tau \sim \pi_\theta(\cdot \mid x; h(\alpha))} [J(\tau)]$, where $J(\tau)$ is the non-differentiable metric (e.g., correctness). To attribute the change in this objective, we first derive its gradient as $\nabla_{h(\alpha)} \mathcal{J}(\alpha) = \mathbb{E}_{\tau} [\sum_{t} \nabla_{h(\alpha)} \log \pi_\theta(a_t \mid s_t; h(\alpha)) A^\pi_t]$. 
Since $h(\alpha) = h' + \alpha(h - h')$, we have $\frac{d h}{d \alpha} = h - h'$.
By the chain rule,
\begin{equation}
    \frac{d\mathcal{J}(\alpha)}{d\alpha}
= \bigl\langle \nabla_{h(\alpha)} \mathcal{J}(\alpha), \; \frac{d h(\alpha)}{d\alpha} \bigr\rangle
= \bigl\langle \nabla_{h(\alpha)} \mathcal{J}(\alpha), \; h - h' \bigr\rangle.
\end{equation}
This is a scalar function of $\alpha$.
The inner product is a sum over the representation dimensions $i = 1,\ldots,d$,
\begin{equation}
    \bigl\langle \nabla_{h(\alpha)} \log\pi_\theta(\tau \mid h(\alpha)), \; h - h' \bigr\rangle
= \sum_{i=1}^d (h_i - h'_i) \nabla_{h_i(\alpha)} \log\pi_\theta(\tau \mid h(\alpha)),
\end{equation}
Next, we apply the fundamental theorem of calculus to the scalar function $\mathcal{J}(\alpha)$. The total difference in performance $\mathcal{J}(1) - \mathcal{J}(0)$ is given by the integral of gradients along the path:
\begin{equation}
\begin{aligned}
    \mathcal{J}(1) - \mathcal{J}(0) &= \int_0^1 \frac{d \mathcal{J}(\alpha)}{d \alpha} \, d\alpha = \int_0^1 \langle \nabla_{h(\alpha)} \mathcal{J}(\alpha), h - h' \rangle \, d\alpha \\
    &= \sum_{i=1}^d (h_i - h'_i) \int_0^1 \mathbb{E}_{\tau \sim \pi_\theta(\cdot \mid h(\alpha))} \left[ \nabla_{h_i(\alpha)} \log \pi_\theta(\tau \mid h(\alpha)) A^\pi(\tau) \right] \, d\alpha.
\end{aligned}
\end{equation}
This derivation confirms that our attribution score $\text{IPG}_i$ represents the functional contribution of the $i$-th component to the expected return, satisfying $\sum_i \text{IPG}_i = \mathbb{E}[J]_{\text{original}} - \mathbb{E}[J]_{\text{baseline}}$. 

\section{Implementation Details}
\label{Sec:appimplementation}
The reproducible code will be released upon the acceptance of this paper.
\subsection{SAE Training}
\label{Sec:apptrainsae}
\textbf{k-Sparse Autoencoder Formulation. }
A k-Sparse Autoencoder (k-SAE)~\citep{gao2025scaling,makhzani2013k} consists of an encoder and a decoder which aim to reconstruct model activations while enforcing sparsity in the feature space. 
The encoding and decoding processes are shown as 

\begin{equation}
    \label{Eq:encode}
    \mathbf{f} = [f_1, f_2, \ldots, f_m] 
    = \text{TopK}(\bW_{\text{enc}}(\mathbf{z} - \mathbf{b}_{\text{pre}}))~, \quad \hat{\mathbf{z}} =\bW_{\text{dec}}\mathbf{f} + \mathbf{b}_{\text{pre}} + \epsilon(\mathbf{z}),
\end{equation}
where $\mathbf{f}$ has non-zero elements $|\mathbf{f}|_0 = k$ ($k < m$), $\mathbf{W}_{\text{enc}} \in \mathbb{R}^{m \times n}$ is the encoder weight, $\mathbf{b}_{\text{pre}} \in \mathbb{R}^{n}$ is the bias term, and $\text{TopK}(\cdot)$ retains the top $k$ largest values while zeroing others.
$\bW_{\text{dec}} \in \mathbb{R}^{n \times m}$ is the decoder weight. Besides, we also add an SAE error term $\mathbf{\epsilon}(\mathbf{z}) = \mathbf{z}-\hat{\mathbf{z}} \in \mathbb{R}^n$ to enable localized modification on residual stream.

\textbf{Dataset for training k-SAEs. }In this section, we provide details about training the k-SAE. Followed by \citep{galichin2025icoveredbaseshere}, we train the SAE on the activations of the model using the full \textit{OPENTHOUGHT-114K} dataset \citep{openthoughts}, which is composed of high quality reasoning trace generated by DEEPSEEK-R1 including math, science, code, etc. To elicit consistent chain-of-thought style activations when harvesting activations, we apply the following chat template:

\begin{verbatim}
<system>Please reason step by step and put your answer within \boxed{}.

</system><|user|>{question}<|assistant|>{deepseek_reasoning}
{deepseek_solution}
\end{verbatim}


\textbf{Training K-SAE on Hidden Space. }
As shown in~\cref{Eq:encode}, the k-SAE contains an encoder $\mathbf{W}_\text{enc}$ and a decoder $\bW_{\text{dec}}$ initialized with the same parameters. The reconstruction loss we use is a standard mean squared error (MSE) loss defined as  
\begin{equation}
    \label{eq:saeloss}
    \mathcal{L}(\mathbf{z}) = \|\mathbf{z} - \hat{\mathbf{z}}\|_2^2
\end{equation}
where $\mathbf{z}$ is the original hidden state vector and $\hat{\mathbf{z}}$ denotes its reconstruction (excluding the residual error term). The dimension of the sparse representation space is $24576$ for Qwen2.5-Math-1.5B-Instruct \citep{yang2024qwen25mathtechnicalreportmathematical} and $65536$ for Llama-3.1-8B-Instruct~\citep{grattafiori2024llama3herdmodels} corresponding to an expansion factor of 16. We train k-SAE on a single NVIDIA H20 GPU, taking approximately 7 hours for Qwen and 30 hours for Llama.

\begin{figure}[htbp]
    \centering
    \begin{subfigure}[b]{0.48\textwidth}
    
        \centering
        \includegraphics[width=\linewidth]{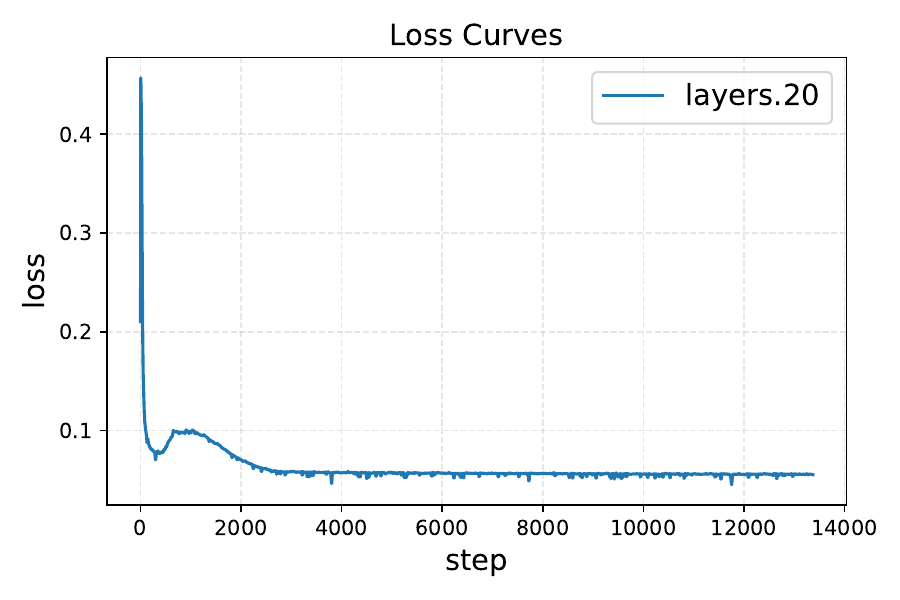} 
        \caption{Training Curve for Qwen2.5-Math-1.5B-Instruct}
    \end{subfigure}
    \hfill
    \begin{subfigure}[b]{0.48\textwidth}
        \centering
        \includegraphics[width=\linewidth]{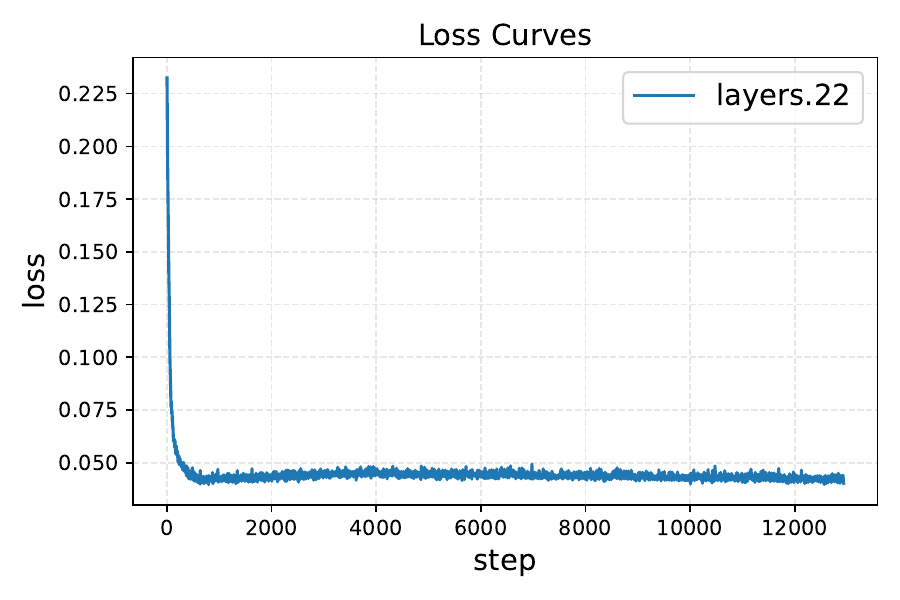}
        \caption{Training curve for Llama-3.1-8B-Instruct}
    \end{subfigure}
    \caption{K-SAE Training Curve}
    \label{fig:loss}
    \vspace{-1.5pt}
\end{figure}

\textbf{Training Results. } We present the details about the training of k-SAEs, including hyper-parameters and learning curves. We use the learning rate of 0.005 with batch sizes of 32, context length of 2048 tokens. For the training curve, we leverage Fraction of Variance Unexplained (FVU) \citep{makhzani2013k}, which is a related metric of interest, measuring the total amount of the original activation that is not ``explained'' or reconstructed well by k-SAE. 
FVU is formally defined as 
\begin{equation}
    FVU=\frac{ \mathcal{L}(\mathbf{z})}{\text{var}[\mathbf{z}]}
\end{equation}
where $\mathbf{z}$ is the hidden state, $\mathcal{L}(\mathbf{z})$ is defined in~\cref{eq:saeloss} and var represents the variance of $\mathbf{z}$. 
A lower FVU indicates better reconstruction performance since more original activation is captured by the k-SAE model.The training curves in~\cref{fig:loss} show that the k-SAE performs well on the FVU metric.

\textbf{Discussion: The Role of SAE Error Node. }
While the k-SAE can effectively reconstruct the hidden state $\hat{\mathbf{h}}$, residual gaps inevitably remain. Recent studies \citep{lindsey2025biology,marks2025sparse} highlight that explicitly modeling error nodes is essential to fill these gaps, as it gives
a principled decomposition of model behaviors into contributions from interpretable features and error components not yet captured by our SAEs.
In our setting, we retain the SAE error term to preserve the reliability of our downstream intervention, thus ensuring that our modification is incremental without degrading the model's performance.

\subsection{IPG Attribution Details}
\label{sec:appipgdtails}
\textbf{Hyperparameter for Attribution. }

This section summarizes the hyperparameter used for the attribution method throughout our experiments. 
Unless otherwise specified, the same configuration is applied across all datasets and models.
The attribution is on one single NVIDIA A40 GPU for 2 hours for Qwen2.5-Math-1.5B-Instruct~\citep{yang2024qwen25mathtechnicalreportmathematical} and one single NVIDIA H20 GPU for Llama3.1-8B-Instruct~\citep{grattafiori2024llama3herdmodels}
\begin{table}[h]
\centering
\caption{IPG-GRPO Hyperparameter Settings.}
\label{tab:attr_hyperparams}
\begin{tabular}{ll}
\toprule
\multicolumn{2}{l}{\textbf{Attribution Configuration}} \\
\midrule
Path integral steps & 10 \\
GRPO group size & 3,5 \\

\midrule
\multicolumn{2}{l}{\textbf{Decoding Strategy}} \\
\midrule
Top-$p$ & 0.8 \\
Temperature & 0.6 \\
Max Generation Tokens & 512,2048,8192 \\
\bottomrule
\end{tabular}
\end{table}

\subsection{Fine-grained Reasoning Mechanism Discovery}

\subsubsection{Process-level Granularity }
We analyze reasoning behavior at the process level by characterizing how internal components are engaged over the course of multi-step inference.

\paragraph{Activation Frequency.}
For a given component (neuron or SAE feature) and a reasoning segment, we define the \emph{activation frequency} as the proportion of tokens within the segment for which the component is active.
Formally, let $\mathcal{T}$ denote the set of tokens belonging to a reasoning segment, and let $\mathbb{I} (f_t > 0)$ be an indicator function that equals 1 if the feature activation at token $t$ is non-zero (i.e., selected by the SAE), and 0 otherwise.
The activation frequency is defined as
\begin{equation}
\mathrm{Freq}(f) \;=\; \frac{1}{|\mathcal{T}|} \sum_{t \in \mathcal{T}} \mathbb{I}(f_t > 0).
\end{equation}
This normalization by the segment length ensures that activation frequency is not confounded by variations in token count across reasoning steps or problems.
All statistics are first computed per example and then averaged across the dataset.

\paragraph{Reasoning Step Delimitation.}
To obtain discrete reasoning steps, we segment each chain-of-thought by line breaks, treating consecutive text blocks separated by the delimiter ``\texttt{\textbackslash n\textbackslash n}'' as individual reasoning steps.
This heuristic follows common formatting conventions in chain-of-thought reasoning, where each paragraph typically corresponds to a coherent sub-step such as problem decomposition, intermediate calculation, or verification.

\paragraph{Step Deciles.}
As different problems may contain varying numbers of reasoning steps, we normalize the reasoning progress of each example to the interval $[0,1]$.
Specifically, for a chain-of-thought consisting of $S$ reasoning steps, the $i$-th step is assigned a normalized progress value $(i-1)/(S-1)$.
We then discretize the normalized reasoning progress into 8 equal-width bins, referred to as \emph{reasoning step deciles}.
For each decile, activation frequency statistics are aggregated across all reasoning steps whose normalized progress falls within the corresponding bin.
This procedure enables alignment and comparison of temporal activation patterns across examples with different reasoning lengths.

\section{Experiment Settings}
In this section, we provide details on the experiment settings (Section~\ref{Sec:expsetup}), including our evaluation pipeline and the baseline implementation.
\subsection{Evaluation Settings}
For all benchmarks, we employ the lm-evaluation-harness framework \citep{eval-harness} as the evaluation tool. For Qwen2.5-Math-1.5B-Instruct~\citep{yang2024qwen25mathtechnicalreportmathematical}, we set the maximum tokens to 2048, while for Llama-3.1-8B-Instruct~\citep{grattafiori2024llama3herdmodels}, we set the maximum new tokens to 8192. Followed by prior work~\citep{dutta2024how} , interventions with IPG are applied to the mid-to-late transformer blocks, specifically layers 20--26 for Qwen and layers 22--28 for Llama.
To avoid randomness and ensure reproducibility, we set ``do sample'' to false, which is equivalent to greedy decoding strategy (temperature set as 0). 
Accordingly, we employ Accuracy@1 as our metric. 

For CMA proxy, we repeat the evaluation with five random seeds on both
Qwen2.5-Math-1.5B-Instruct and DeepSeek-R1-Distill-Qwen-1.5B.
The latter provides a longer-trajectory setting, with approximately
4,000 generated tokens on average on MATH-500, compared with
approximately 550 for Qwen.
We construct a group-based mediation proxy
by randomly sampling neuron groups, zero-ablating their activations
at a target layer during generation, and selecting groups with high drop of reasoning ability measured by externally verified correctness.

\label{Sec:appevalmetric}
\subsection{Baselines}
\label{Sec:baselines}
For the baseline, all provided results are fair, including searching for their best configurations according to their presented hyperparameter selection rules. 
\paragraph{Random. }
For comparison, we construct a random baseline by uniformly and randomly selecting neurons within the hidden states $\bh$ and applying interventions based on these neurons with the same number of our \method. 
The final results of this baseline are averaged across multiple random seeds.
\paragraph{Activation. }
Internal neuron activations contain meaningful information. 
Additionally, \citep{dai-etal-2022-knowledge} demonstrates the ability to edit internal neurons to influence neural network performance. 
Building on this, we adopt a similar approach, performing intervention directly on the original activations in the hidden state. 
For a fair comparison, we constrain our intervention to the residual stream of a language model.
\paragraph{Reasoning Neuron. } As described in~\citep{rai-yao-2024-investigation}, the focus is on analyzing activation patterns, such as average activation values in the feed-forward (FF) layer, to identify reasoning neurons and interpret them using predefined concepts $C$, such as arithmetic operations $C_{\text{add}}$ with key tokens like ${\text{``add''}, \text{``+''}}$. 
We extend this approach to operate on the residual stream of transformer blocks~\citep{vaswani2017attention}. 
By leveraging all provided concepts, we identify related neurons and use them for intervention.
\paragraph{Control Vector. } Introduced by~\citep{hjer2025improving}, it employs representation engineering by extracting model activations from the residual stream of a large language model (LLM) during a reasoning task. 
These activations are used to derive a control vector for inference-time intervention.
The control vector is constructed from positive and negative pairs: positive pairs consist of Chain-of-Thought (CoT) prompts where the model produces correct outputs, while negative pairs comprise 75 random character strings sampled from the alphabet (A–Z). Then PCA is applied and the first principal component in the control direction is treated as the most reasoning-related direction. Following the open-source repository, we use 30\% of the samples from each dataset to compute the control vector.

\paragraph{SAE-Reasoning. } As introduced by 
~\citep{galichin2025icoveredbaseshere}, it employs Sparse Autoencoders (SAEs)~\citep{makhzani2013k,gao2025scaling} and learn a sparse decomposition of latent representations of a neural network into interpretable features, to identify features that drive reasoning in the DeepSeek-R1~\citep{guo2025deepseek} series of models. 
They extract candidate ``reasoning feature'' from SAE representations, that based on specific tokens such as $\{\text{``wait''},\text{``but''}\}$, associating feature activation with these tokens. For each feature,  \emph{reasonscore} is computed based on its feature activation on the \emph{OPENTHOUGHT-114K} \citep{openthoughts}. After that, following the releasing implementation, we intervene the feature with top \emph{reasonscore}.

\section{More results}
In this part, we present more experiments and studies for better illustration of our \method. 
\subsection{Sensitivity Analysis}
\subsubsection{K-SAE hyperparameter}
\label{Sec:appsensitiveksae}

We adjust different $k$ values to evaluate the effects of the k-SAEs~\citep{gao2025scaling} on reasoning behavior control as shown in Table~\ref{table:ablation_topk}.
The results are based on Qwen2.5-Math-1.5B-Instruct~\citep{yang2024qwen25mathtechnicalreportmathematical} layer 20.

    
\begin{table}[htbp]
\centering
\caption{Investigation of  the impact of varying TopK parameters in k-SAE~\citep{gao2024kvcachesteering} on controlling reasoning behavior in Qwen2.5-Math-1.5B-Instruct~\citep{yang2024qwen25mathtechnicalreportmathematical} for GSM8K~\citep{cobbe2021trainingverifierssolvemath}. TopK refers to preserving the top $k$ features while deactivating others in the feature space, as described in Section~\ref{Sec:intervene}.}
\resizebox{0.3\linewidth}{!}{
 \begin{tabular}{@{}cc@{}}
 \hline
 \multirow{2}{*}{$k$} 
 & \multicolumn{1}{c}{\textbf{GSM8K}} \\ 
 & \textbf{Acc. (Enhance)~$\uparrow$} \\
 \hline
 Original & 82.41 \\
 8 & {84.38}   \\
  \textbf{32} & \textbf{84.76} \\
 64 & 83.70  \\
 \hline
 \end{tabular}
}
\label{table:ablation_topk}
\end{table}
In Table~\ref{table:ablation_topk}, as the TopK parameter $k$ increases, the ability to control reasoning behavior, including both enhancing and suppressing, either persists or begins to diminish. 
We demonstrate that our results are not heavily dependent on the choice of Sparse Autoencoders (SAEs). Instead, our focus is on the accurate identification and intervention of reasoning features. Consequently, we select $k=32$ for the k-SAE~\citep{gao2025scaling} across all experiments.

\subsubsection{\method Attribution Hyperparameter}
\label{Sec:appattrhyper}
We conduct experiment on the effects of different number $p$ of components identified, as stated in~\cref{Sec:ipgattribution}.
\begin{table}[htbp]
\centering
\caption{Evaluation on the impact of varying the top-$p$ number of neurons in~\ref{Sec:intervene} in Qwen2.5-Math-1.5B-Instruct~\citep{yang2024qwen25mathtechnicalreportmathematical} for GSM8K~\citep{cobbe2021trainingverifierssolvemath}.This evaluation is based on our k-SAE implementation~\citep{gao2025scaling} with a fixed $k=32$.}
\resizebox{0.3\linewidth}{!}{
 \begin{tabular}{@{}cc@{}}
 \hline
 \multirow{2}{*}{$p$} 
 & \multicolumn{1}{c}{\textbf{GSM8K}} \\ 
 & \textbf{Acc. (Enhance)~$\uparrow$} \\ 
 \hline
 Original & 82.41  \\
 10 & 82.94  \\
  20 & 83.85  \\
 30 & 83.78 \\
 40 & {83.85} \\
 50 & \textbf{84.60}\\
 60 & 84.15\\
 \hline
 \end{tabular}
}
\label{Tab:sensiigtopp}
\end{table}
In Table~\ref{Tab:sensiigtopp}, as the number of top-$p$ neurons increases, the control over reasoning behavior becomes more effective. 
However, beyond a certain point, adding more neurons does not yield further improvements. 
This is because reasoning-related concepts may be distributed across multiple neurons, and the top neurons already exert significant control, while additional neurons lack substantial reasoning-related functionality.

We also conduct sensitivity analysis on our choice of batch size $M$. We perform an ablation on the number of samples used to compute the \method directions, varying it from 1 to 500. All other settings follow the main experimental protocol (Table~\ref{tab:batch-ablation}): Qwen2.5-Math-1.5B-Instruct~\citep{team2024qwq} evaluated on GSM8K~\citep{cobbe2021trainingverifierssolvemath} under greedy decoding. ``Enhance'' reports accuracy after strengthening the identified reasoning components (higher is better), while ``Suppress'' reports accuracy after weakening them (lower is better). 
The original accuracy of the model is 82.41\%.

When $M$ is very small ($\leq 10$), the gradient estimates exhibit high variance, which degrades component identification and leads to weak performance.  
As it increases to $\geq 20$, variance decreases rapidly, yielding smoother and more reliable gradient signals. This stabilizes both enhancement and suppression, with suppression improving dramatically.

\begin{table}[tb]
\centering
\small
\caption{Effect of batch size (\# samples used to compute \method  directions) on steering performance. Model: Qwen2.5-Math-1.5B-Instruct on GSM8K~\citep{cobbe2021trainingverifierssolvemath}. Original accuracy: 82.41\%.}
\begin{tabular}{@{}cc@{}}
 \hline
 \multirow{2}{*}{\# Samples} 
 & \multicolumn{1}{c}{\textbf{GSM8K}} \\ 
 & \textbf{Acc. (Enhance)~$\uparrow$}  \\ 
 \hline
1     & 82.79  \\
10    & 83.17 \\
20    & 83.17  \\
50    & 83.54 \\
100   & 83.39  \\
500  & \textbf{83.55}  \\
\bottomrule
\end{tabular}
\label{tab:batch-ablation}
\end{table}

\subsubsection{Reward Signals}

\begin{table}[htbp]
\centering
\setlength{\tabcolsep}{5pt} 
\caption{Performance of \method in enhancing accuracy across reasoning datasets for Qwen2.5-Math-1.5B-Instruct~\citep{yang2024qwen25mathtechnicalreportmathematical} under different reward signals.}
\begin{tabular}{l|l|cccc}
\toprule
\textbf{Model} & \textbf{Reward Method}
    & \makecell{\textbf{GSM8K}}
    & \makecell{\textbf{MATH-500}}
    & \makecell{\textbf{AIME-2024}}
    & \makecell{\textbf{GPQA}\\\textbf{Diamond}} \\
\midrule
\multirow{2}{*}{Qwen2.5-Math-1.5B}
   & Rule Function   & 84.15 & 64.20 & 20.00 & 26.77 \\
   & Reward Model & 84.60 & 65.80 & 20.00 & 35.35 \\
\bottomrule
\end{tabular}
\label{tab:reward_signals_usage}
\end{table}
We report the performance of \method in improving accuracy across reasoning datasets for Qwen2.5-Math-1.5B-Instruct~\citep{yang2024qwen25mathtechnicalreportmathematical} using different reward signals, including rule-based and reward model-based signals, as detailed in Appendix~\ref{Sec:appalgvariants}. 
The scaling intervention (see Section~\ref{Sec:intervene}) is applied at the 24th layer. 
As shown in Table~\ref{tab:reward_signals_usage}, reward model-based signals outperform rule-based signals in most datasets, suggesting that reward models capture richer information or better represent reasoning outcomes, thereby enhancing \method's effectiveness.

\label{Sec:apprewardsig}

\subsubsection{Baseline Choice.}
\label{Sec:appsensibaseline}
To further verify the robustness of \method, we conduct additional experiments on different choices of the baseline hidden state $h'$ using Qwen-Math-2.5-1.5B-Instruct~\citep{team2024qwq} on the GSM8K~\citep{cobbe2021trainingverifierssolvemath} dataset. 
We report accuracy under greedy decoding as described in Section~\ref{Sec:expsetup}.
The original accuracy of the model is 82.41\%.
We compare three baseline strategies:
(i) the zero vector,
(ii) the mean hidden state computed over the training set,
and (iii) random Gaussian noise $\mathcal{N}(0, I)$ as an explicit non-informative reference.

Results are shown in Table~\ref{tab:baseline-ablation}.
Using the zero vector or the mean hidden state yields nearly identical enhancement performance (84.45\% and 84.60\%, respectively) and both enable almost perfect suppression (10.69\% and 21.15\%).
In contrast, random noise performs considerably worse on suppression (78.16\%), confirming that it does not constitute a meaningful baseline representation.
These findings demonstrate that \method is highly robust to the specific choice of $h'$ as long as it represents a reasonable direction; arbitrary noise fails to serve this role effectively.

\begin{table}[tb]
\centering
\small
\caption{Sensitivity analysis on the choice of baseline representation $h'$. 
The model is Qwen-Math-2.5-1.5B-Instruct~\citep{team2024qwq} evaluated on GSM8K~\citep{cobbe2021trainingverifierssolvemath}. 
Original accuracy before any intervention: 82.41\%. }
\begin{tabular}{@{}cc@{}}
 \hline
 \multirow{2}{*}{$\bh'$} 
 & \multicolumn{1}{c}{\textbf{GSM8K}} \\ 
 & \textbf{Acc. (Enhance)~$\uparrow$}  \\
 \hline
Zero vector              & \textbf{84.60}                                                   \\
Mean hidden state        & {84.45}                                                     \\
Random noise             & 84.00                                                            \\
\bottomrule
\end{tabular}
\label{tab:baseline-ablation}
\end{table}

\subsection{Controlling Reasoning Length}
\label{Sec:appresoninglength}

Beyond identifying reasoning components using accuracy-based reward signals, we explore whether neurons controlling the length of reasoning can be identified. 
This is achieved by modifying the rule function (see Appendix~\ref{Sec:appalgvariants}) to prioritize sensitivity to reasoning length rather than accuracy. 
Specifically, the rule function is redefined to reward outputs based on their generated length, encouraging the identification of neurons that influence the verbosity or conciseness of reasoning.

The modified signal function, $J(\cdot)$, is defined as a function of the generated text length, $|\tau|$, measured as the number of tokens in the output, $\tau$ is the generated trajectory. 
The reward increases with length to promote longer reasoning chains or decreases to favor conciseness, depending on the desired behavior. 
The rule function is given by
\begin{equation}
J({\tau}) = f(|\tau|) \;,
\end{equation}
where $f(\cdot)$ can be any monotonically increasing function.
For example, we can have a quadratic function $f(L)= \alpha L^2+\beta L + \gamma$
where $\alpha$, $\beta$, and $\gamma$ are tunable parameters that control the reward's sensitivity to length.

As shown in~\cref{fig:lengthcontrol} in~\cref{sec:reasoning_identicontrol}, scaling the top one percent of neurons identified via length-based rewards on DeepSeek-Qwen-1.5B~\citep{guo2025deepseek} leads to a gradual decrease in the model's output length as the scaling factor increases.
In conclusion, these results demonstrate that our \method can be readily extended to capture diverse outcome-aware signals corresponding to specific behaviors, highlighting its flexibility and effectiveness.

\subsection{Validation of Reasoning Components under sampling evaluation.}
\label{sec:appvadiation}
As discussed in \cref{Sec:expsetup}, our method remains effective under sampling‑based evaluation, where multiple (16) candidate solutions are generated.
Table~\ref{tab:appendix_suppressing} reports the accuracy on GSM8K, MATH-500, and AIME-2024 under such a setting.
We compare the original model (without our reasoning‑strength control) against our method using the same sampling budget.
Across all three benchmarks, our method yields consistent improvements.
These results confirm that the reasoning components governed by our approach are beneficial not only in greedy decoding but also when exploration is encouraged through sampling.
\begin{table*}[t]
\small
\centering
\caption{Comparison of \method and baseline methods on pass@8 accuracy across reasoning datasets for Qwen2.5-Math-1.5B-Instruct \citep{yang2024qwen25mathtechnicalreportmathematical} 
}
\vspace{-5pt}
\resizebox{0.5\linewidth}{!}{%
\begin{tabular}{lccc}
\toprule
\textbf{Method} &
\makecell{GSM8K} &
\makecell{MATH-500} &
\makecell{AIME-2024}\\
\midrule
Original & 93.83 & 76.38 & 25.07 \\
\textbf{\method (Ours)}    & \textbf{94.77} & \textbf{77.77} & \textbf{32.61} \\
\bottomrule
\end{tabular}
}\label{tab:appendix_suppressing}
\vspace{-10pt}
\end{table*}

\subsection{Ablation on Policy Gradient}
\label{Sec:appablapg}
In the main paper, we accumulate policy gradients along a continuous path from a baseline activation $\mathbf{h}'$ to the original activation $\mathbf{h}$, yielding \emph{Integrated Policy Gradients} (IPG). 
This path-integral formulation produces baseline-aware attribution scores that reduce gradient noise and improve faithfulness. In this section, we provide an ablation study comparing IPG against \emph{raw policy gradient} (PG) attribution.

\paragraph{Experimental Setup. }We conduct experiments on \textsc{Qwen2.5-Math-1.5B-Instruct}~\citep{yang2024qwen25mathtechnicalreportmathematical} using two mathematical reasoning benchmarks: \textsc{MATH500}~\citep{hendrycks2021measuringmathematicalproblemsolving} and \textsc{AIME2024}~\citep{aime2024}. We report accuracy (\%). 
For each dataset, we evaluate five settings: the original model without intervention, PG-based attribution (with and without SAE), and IPG-based attribution (with and without SAE). Importantly, all experimental settings share identical implementations and hyperparameters; the only difference lies in the attribution signal, i.e., raw policy gradients versus integrated policy gradients.

\paragraph{Results.}
Table~\ref{tab:pg_ablation} summarizes the results. Raw policy gradient attribution provides only modest improvements over the original model, and its effectiveness is limited even when combined with SAE. 
In contrast, IPG consistently yields larger performance gains across both datasets, with or without SAE. Notably, IPG achieves improvements on the more challenging \textsc{AIME2024} benchmark, highlighting its robustness in difficult reasoning scenarios.

\begin{table}[h]
\centering
\caption{Ablation study comparing raw policy gradient (PG) and integrated policy gradient (IPG) attribution on \textsc{Qwen2.5-Math-1.5B-Instruct}. Accuracy (\%) is reported.}
\label{tab:pg_ablation}
\begin{tabular}{lcc}
\toprule
Method & MATH500~$\uparrow$ & AIME2024~$\uparrow$ \\
\midrule
Original & 63.00 & 10.00 \\
\hline
PG & 64.60 & 16.67 \\
\method & \textbf{65.80} & \textbf{20.00} \\
\hline
PG-SAE & 64.40 & 16.67 \\
\method-SAE & \textbf{66.00} & \textbf{20.00} \\
\bottomrule
\end{tabular}
\end{table}

\paragraph{Analysis and Conclusion.}
These results demonstrate that the gains of IPG are not merely due to policy gradient attribution itself, but critically depend on the \emph{integration along the activation path}. By aggregating gradients across activation levels, IPG mitigates local gradient noise and captures sustained, cumulative contributions of internal components. This leads to more reliable identification of influential components and, consequently, more effective intervention. Overall, this ablation confirms that the \emph{integrated} design is a key factor underlying the superior performance of IPG.

\subsection{Comparison with Component-Attribution Baselines}
\label{app:attribution_baselines}

We compare \method with Integrated Gradients (IG), Neuron Shapley, and AttnLRP on Qwen2.5-Math-1.5B-Instruct.
For these baselines, we first sample rollouts from the same model and hold the generated tokens fixed during attribution.
The scalar attribution objective is the sum of the log probabilities of the answer tokens.
The comparison uses the same task-level aggregation, top-neuron selection, and intervention protocol across methods.
All interventions are performed in the neuron space, and evaluation uses greedy decoding.

Table~\ref{tab:attribution_baselines} shows that each baseline improves accuracy over the original model, indicating that conventional attribution methods also identify components useful for reasoning.
\method achieves the highest accuracy across all four benchmarks.
The comparison supports the practical benefit of its outcome-based sequential attribution, although the methods also differ in their attribution objectives.

\begin{table}[t]
\centering
\small
\caption{
Accuracy (\%) under neuron-space enhancement on
Qwen2.5-Math-1.5B-Instruct.
The highest accuracy in each column is shown in \textbf{bold}.
}
\label{tab:attribution_baselines}
\begin{tabular}{lcccc}
\toprule
\textbf{Method} & GSM8K & MATH & AIME & GPQA \\
\midrule
Original       & 82.41 & 63.00 & 10.00 & 24.75 \\
IG             & 82.94 & 64.40 & 13.33 & 29.29 \\
Neuron Shapley & 83.47 & 64.80 & 13.33 & 30.30 \\
AttnLRP        & 83.62 & 64.00 & 16.67 & 28.28 \\
\textbf{\method}
& \textbf{84.61}
& \textbf{65.80}
& \textbf{20.00}
& \textbf{33.33} \\
\bottomrule
\end{tabular}
\end{table}

\subsection{General Reasoning Mechanism Identification}
\label{Sec:appgeneralmechan}
\paragraph{Metric Definition.}
We quantify consistency using the Dice similarity coefficient. 
Formally, let $\mathbf{P_A}$ and $\mathbf{P_B}$ denote the sets of neuron indices identified by a method in datasets $A$ and $B$, respectively. The similarity score is defined as:
\begin{equation}
    \text{Dice}(\mathbf{P_A}, \mathbf{P_B}) = \frac{2|\mathbf{P_A} \cap \mathbf{P_B}|}{|\mathbf{P_A}| + |\mathbf{P_B}|}.
\end{equation}
A higher score indicates that the method identifies a similar set of critical components regardless of the input domain.

\begin{figure}[htbp]
    \centering
    \includegraphics[width=0.5\textwidth]{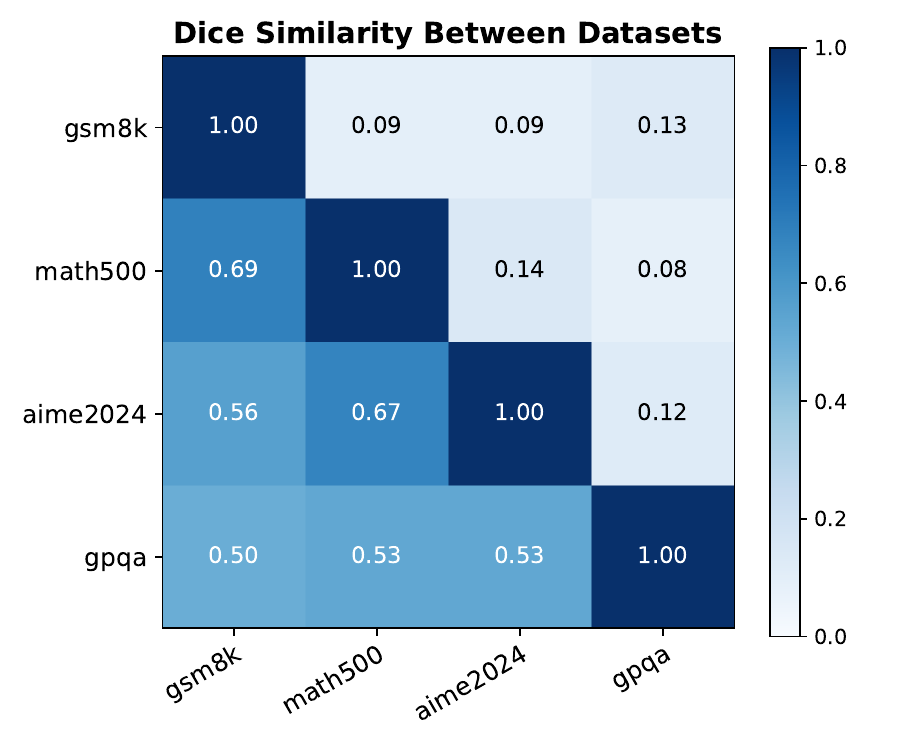}
    \caption{Dice similarity between neuron indices across datasets. We compare neuron indices discovered in 4 different datasets, highlighting shared reasoning structure across tasks, with diagonal showing random baseline. }
    \label{fig:jaccard}
\end{figure}

\paragraph{Quantitative Results.}
\cref{fig:jaccard} presents the pairwise similarity matrix. The matrix is asymmetric to facilitate a direct comparison between our method and a random baseline:
\begin{itemize}
    \item \textbf{Lower Triangular (Proposed \method):} These elements display the overlap of neuron indices discovered by our method. We observe a high degree of consistency across all dataset pairs, with Dice scores ranging from $0.50$ to $0.69$. Notably, the overlap between GSM8K and MATH500 is particularly strong ($0.69$), suggesting that the core reasoning mechanisms for grade-school and competition-level math are highly aligned. Even when comparing vastly different distributions, such as GSM8K and GPQA, the similarity remains robust at $0.50$.
    \item \textbf{Upper Triangular (Random Baseline):} To contextualize these results, the upper triangular elements report the Dice similarity for a random baseline (RND.), where neuron indices are selected uniformly at random. The resulting scores are consistently low, ranging between $0.08$ and $0.14$.
\end{itemize}

\paragraph{Conclusion.}
The significant margin between the lower and upper triangular values confirms that \method captures systematic, task-agnostic reasoning structures. The high overlap scores ($>0.50$) compared to the random baseline ($\approx 0.10$) demonstrate that the identified internal mechanisms are stable and transferable across varying levels of problem complexity.

\subsection{Discussion}

\begin{figure}[htbp]
    \centering
    \begin{subfigure}[b]{0.49\textwidth}
        \centering
        \includegraphics[width=\linewidth]{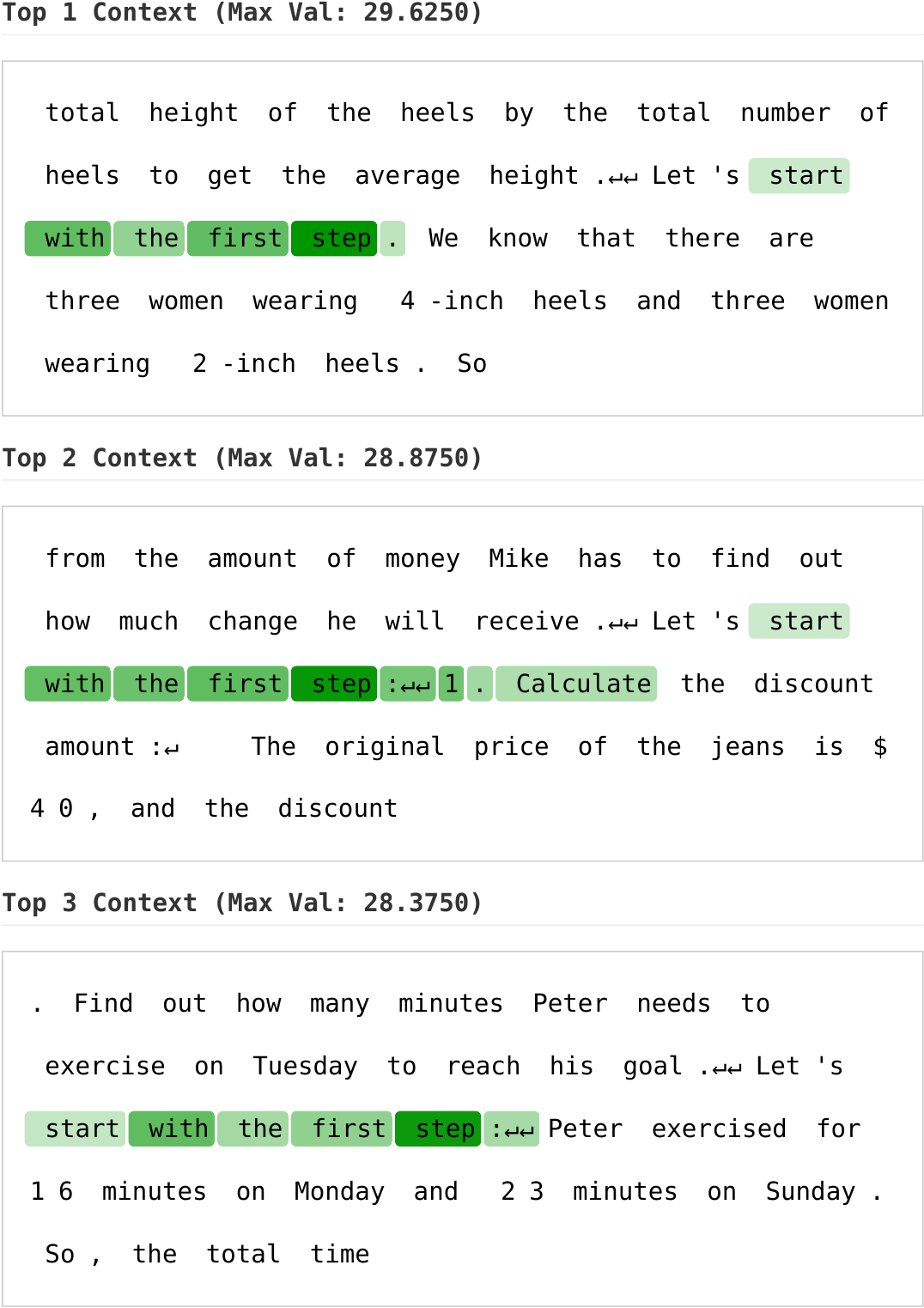}
        \caption*{Feature 486 (layer 20)}
    \end{subfigure}
    \hfill
    \begin{subfigure}[b]{0.49\textwidth}
        \centering
        \includegraphics[width=\linewidth]{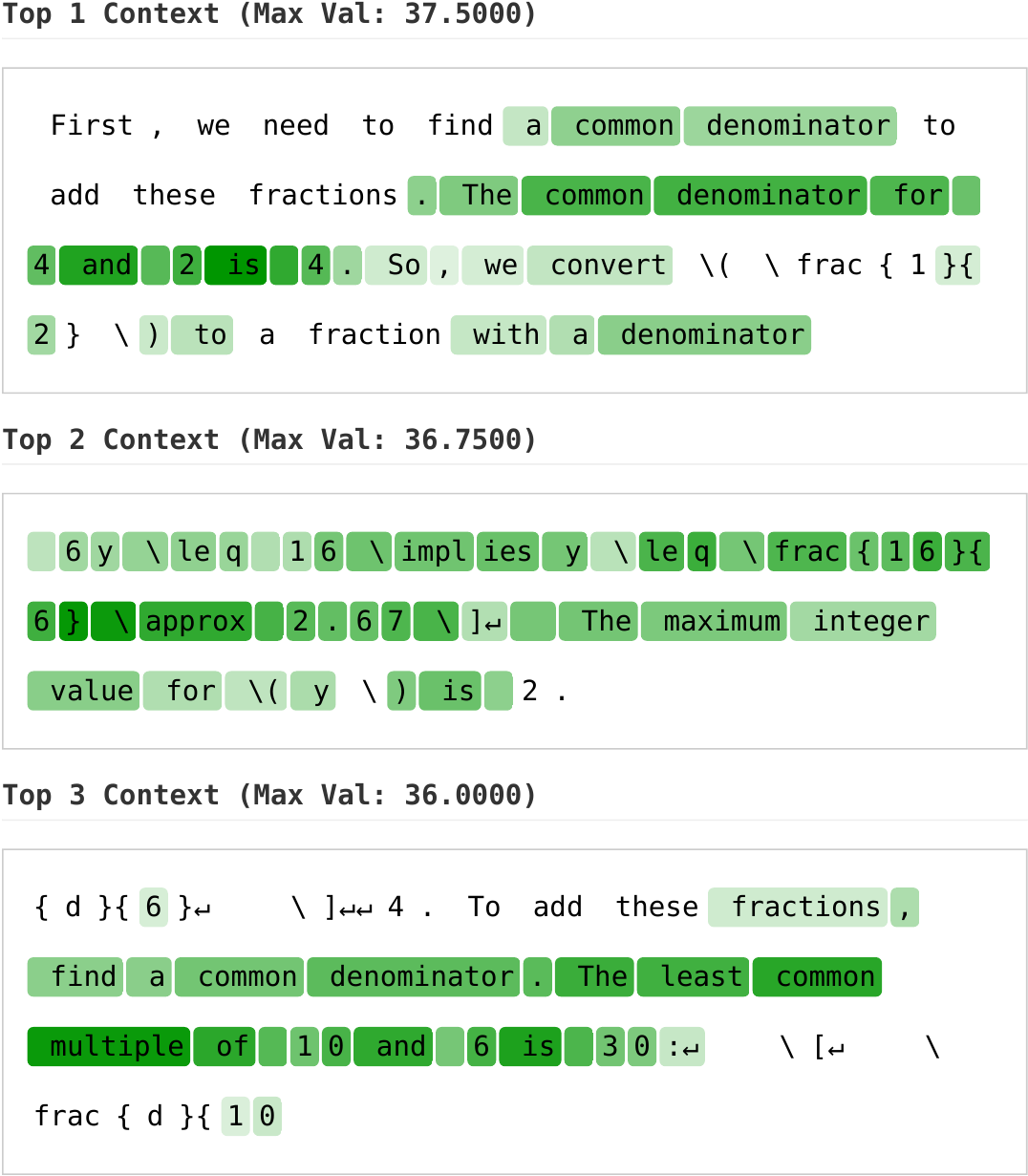}
        \caption*{Feature 513 (layer 20)}
    \end{subfigure}
    \caption{SAE features 486 and 513 from layer 20 of Qwen2.5-Math-1.5B-Instruct, identified by IPG. Tokens receiving higher activation are shown in darker green.}
    \label{fig:sae_features_20}
\end{figure}

\begin{figure}[htbp]
    \centering
    \begin{subfigure}[b]{0.49\textwidth}
        \centering
        \includegraphics[width=\linewidth]{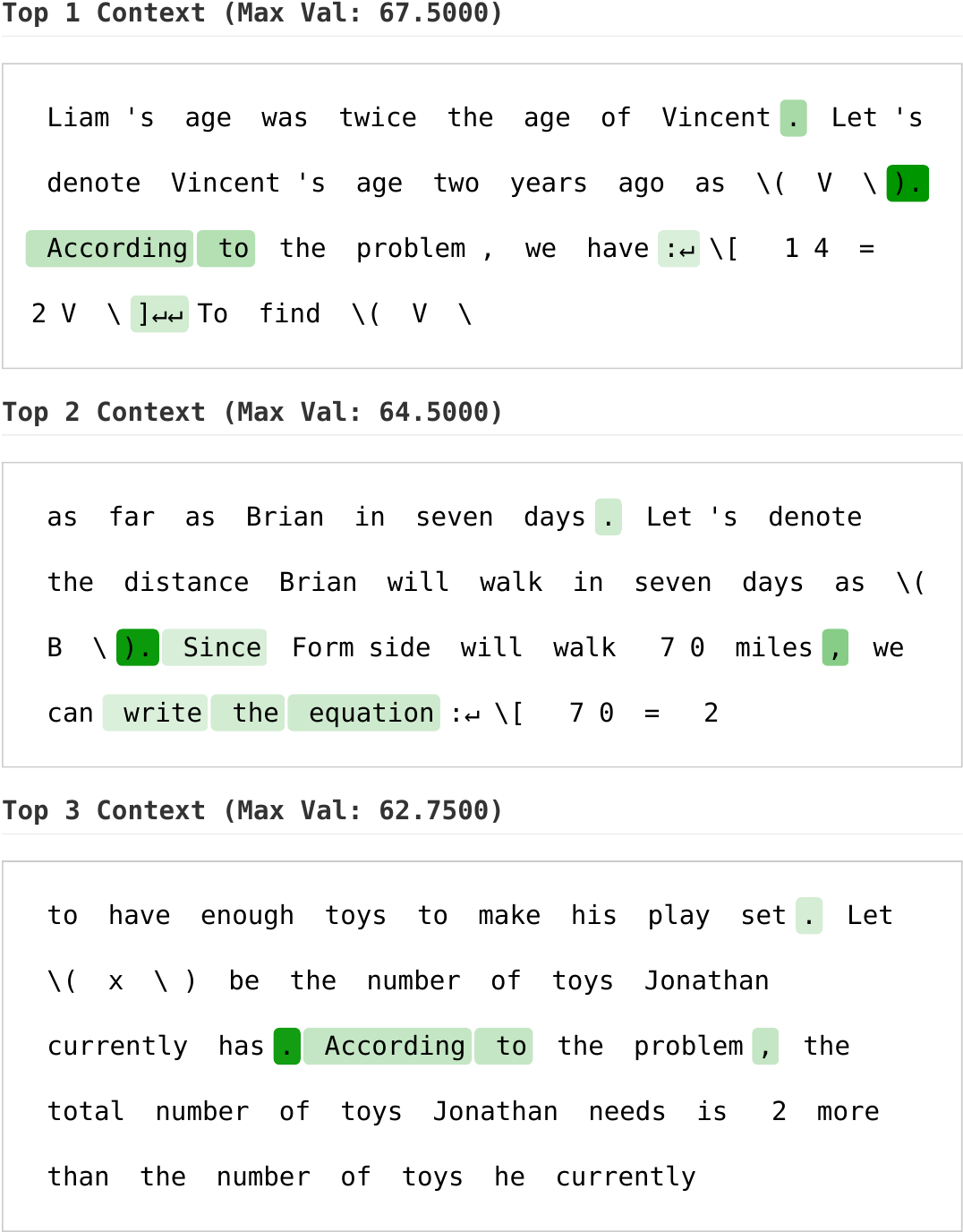}
        \caption*{Feature 12581 (layer 24)}
    \end{subfigure}
    \hfill
    \begin{subfigure}[b]{0.49\textwidth}
        \centering
        \includegraphics[width=\linewidth]{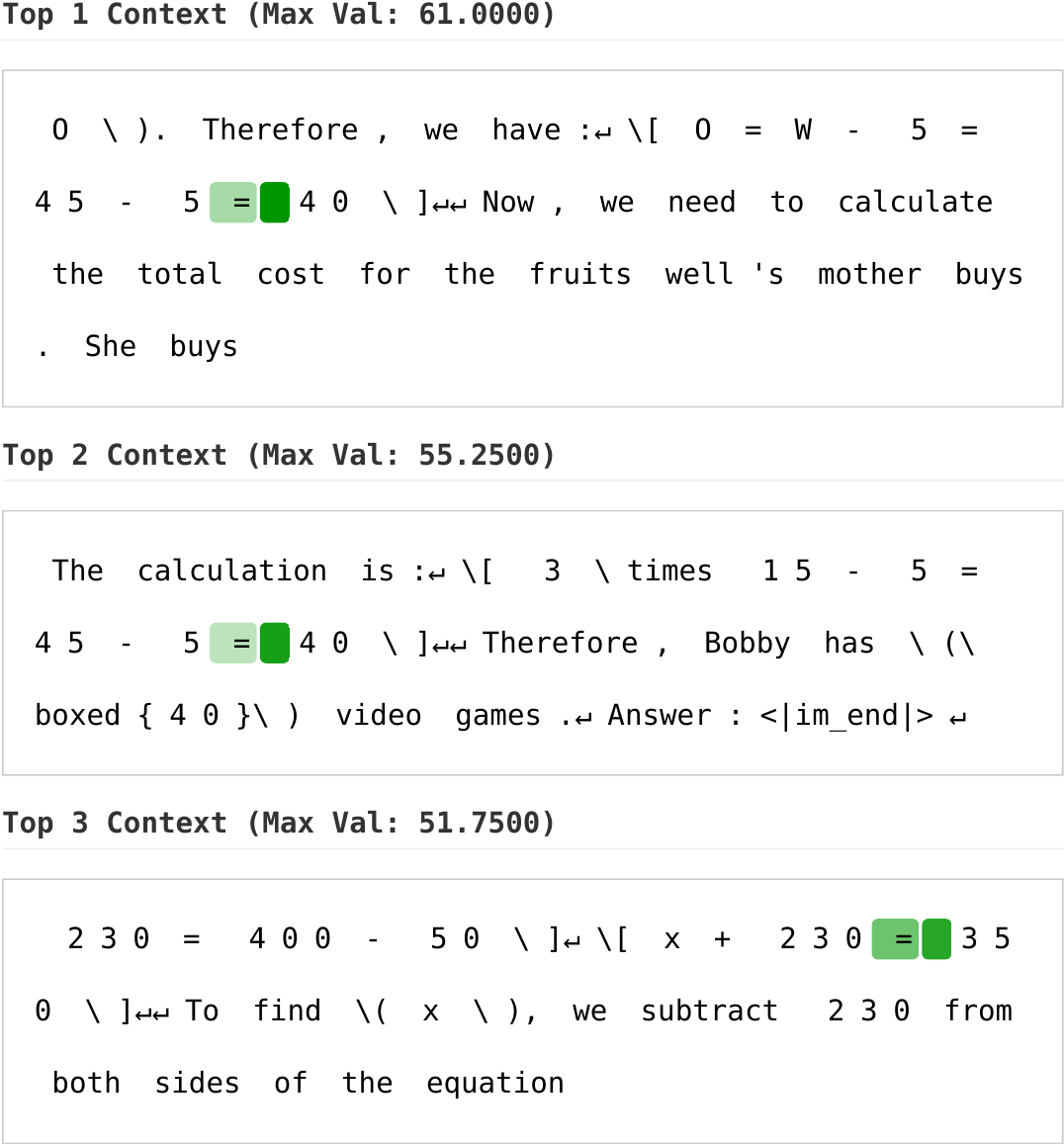}
        \caption*{Feature 13446 (layer 24)}
    \end{subfigure}
    \caption{SAE features 12581 and 13446 from layer 24 of Qwen2.5-Math-1.5B-Instruct, identified by IPG. Tokens receiving higher activation are shown in darker green.}
    \label{fig:sae_features_24}
\end{figure}

\begin{figure}[htbp]
    \centering
    \begin{subfigure}[b]{0.49\textwidth}
        \centering
        \includegraphics[width=\linewidth]{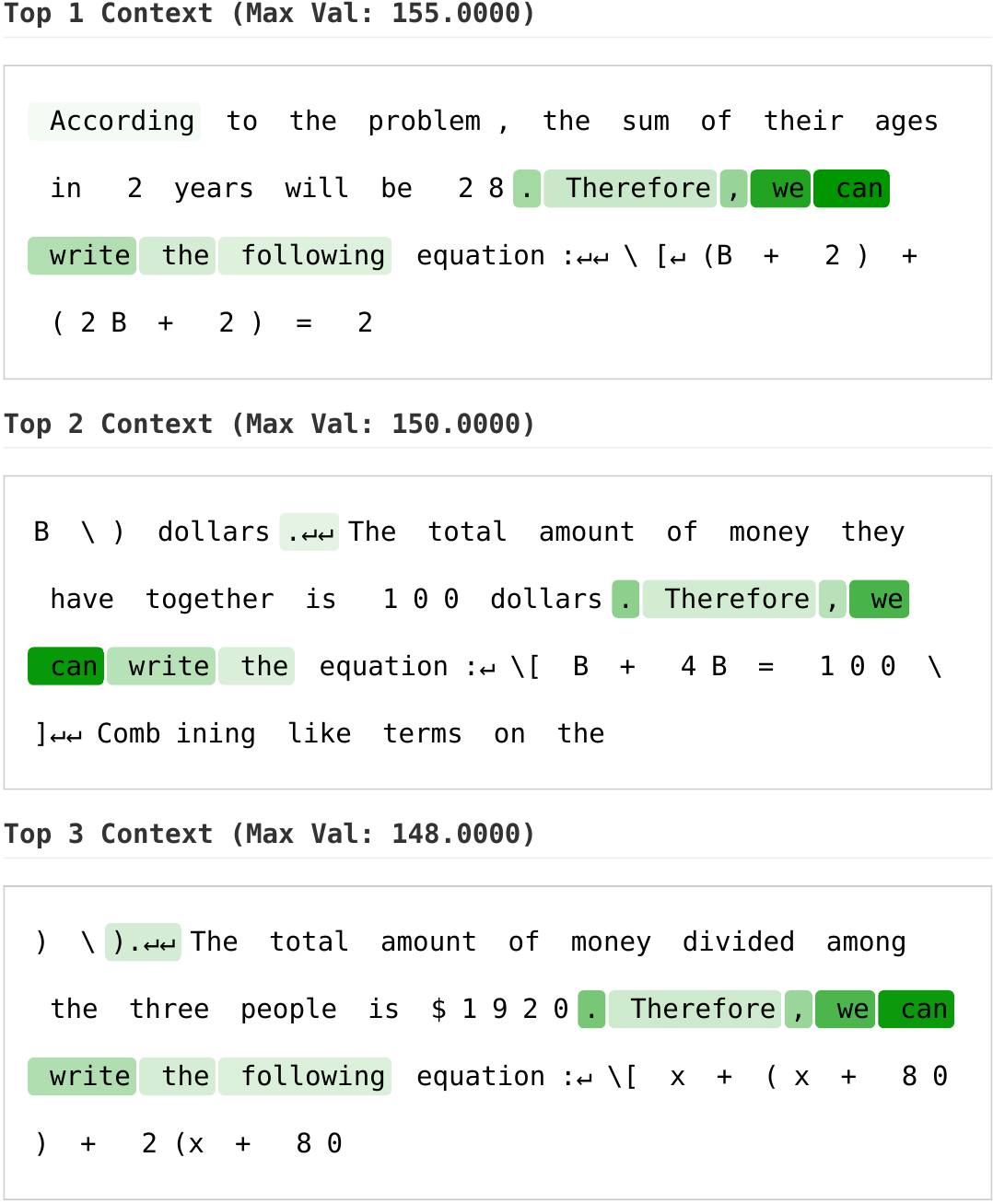}
        \caption*{Feature 13499 (layer 26)}
    \end{subfigure}
    \hfill
    \begin{subfigure}[b]{0.49\textwidth}
        \centering
        \includegraphics[width=\linewidth]{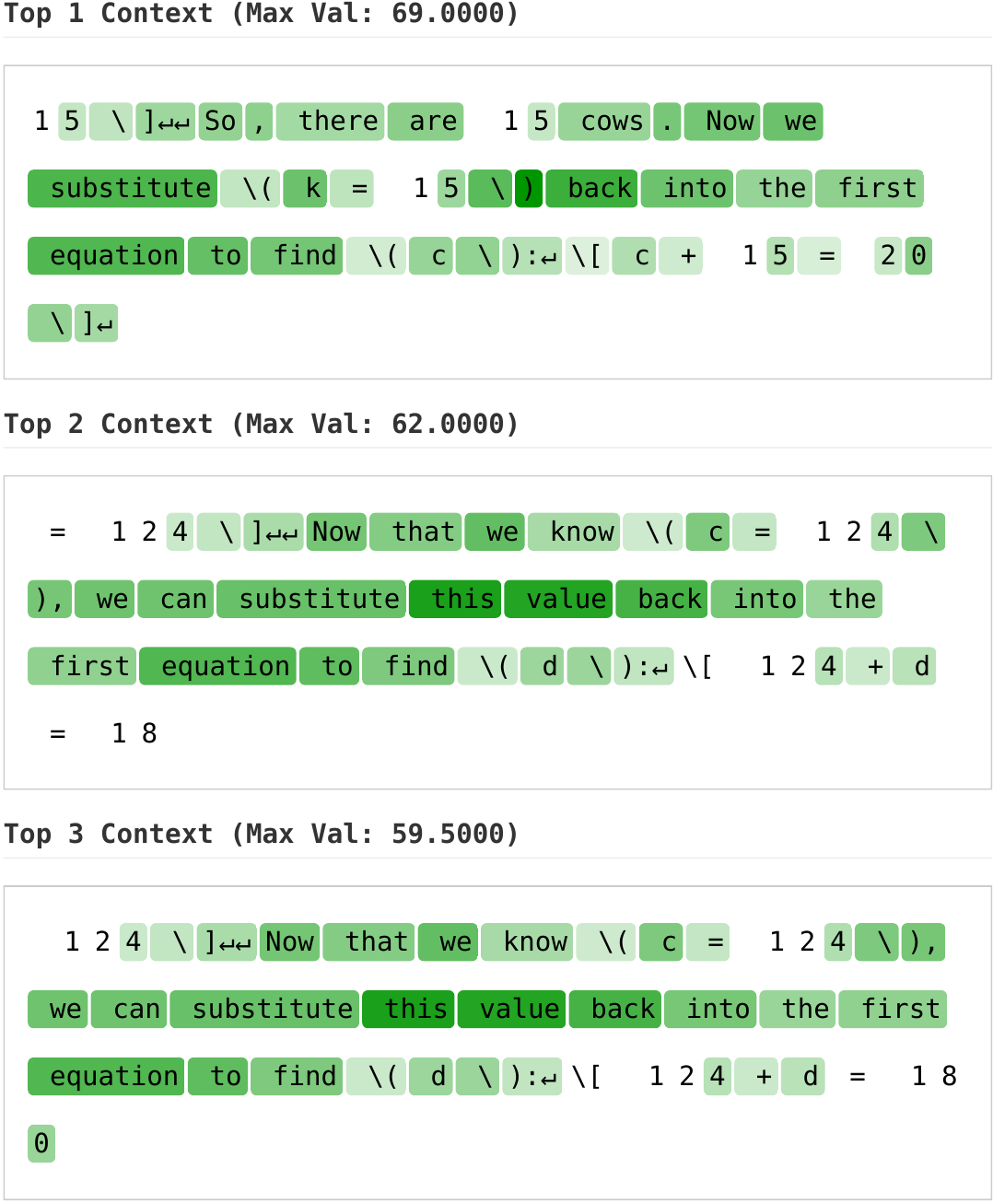}
        \caption*{Feature 24361 (layer 26)}
    \end{subfigure}
    \caption{SAE features 13499 and 24361 from layer 26 of Qwen2.5-Math-1.5B-Instruct, identified by IPG. Tokens receiving higher activation are shown in darker green.}
    \label{fig:sae_features_26}
\end{figure}

\subsubsection{Limitation of IPG}
\label{sec:applimitation}
\textbf{Reward Signal:} Our current implementation of \method relies on relatively preliminary reward definitions, including either a rule-based signal or a lightweight reward model. Such signals can be coarse and fail to capture subtle reasoning improvements. Consequently, they may introduce noise or bias when attributing causal contributions. More sophisticated approaches, such as process reward models (PRMs) or step-level verifiers, could provide finer-grained supervision that better aligns with reasoning quality. Exploring such enhanced reward sources is a promising direction for future work.

\textbf{Selection of Scaling Factor:}
In the present \method implementation, the scaling factor ($\gamma$) is applied uniformly across neurons or features. In spite of its general intervention effects, it neglects more delicate things where different components may require different intervention strengths. A promising future work is to adopt component-specific scaling, allowing each neuron or feature to be steered by an individually optimized factor.

\textbf{Causality-motivation of IPG}
\label{Sec:appdiscausality}
While standard gradient-based attribution method measures local sensitivity, we argue that our \method framework is more \emph{causality-motivated} than correlation-based method (e.g. SAE-R) for two reasons. We list the comparison in Table \ref{tab:ipg_vs_correlation}.

\textit{CMA-style Attribution: }
    \method is conceptually aligned with Causal Mediation Analysis \citep{DBLP:journals/corr/abs-2408-01416}. Inspired by CMA, we treat neuron/SAE feature sets as high-dimensional mediators and quantify their contributions to the final reasoning outcome. Instead of computationally prohibitive brute-force patching, IPG uses path-integrated gradients to propose high-quality candidates for causal mediators.
    
\textit{Causal Validation via Intervention: }
    We do not rely on attribution alone. The causal link is established in Section \ref{Sec:intervene}. We empirically validate causal roles of identified components underlying the reasoning mechanism subsequently in Section \ref{sec:reasoning_identicontrol}.

\begin{table}[t]
\centering
\caption{What IPG Provides Compared to Correlation-Based Methods}
\label{tab:ipg_vs_correlation}
\begin{tabular}{lcc}
\toprule
\textbf{Aspect} 
& \textbf{Correlation-Based (e.g., SAE-R)} 
& \textbf{IPG} \\
\midrule
Signal source 
& Text pattern co-occurrence 
& Reasoning outcome (accuracy/length) \\

Attribution target 
& Activates with reasoning tokens
& Influences final outcome \\

Temporal scope 
& Local / point-wise 
& Long-horizon (full trajectory) \\
\bottomrule
\end{tabular}
\end{table}

\subsection{Case Studies}
\label{Sec:appcasestudy}

\subsubsection{Component-Level Disentanglement}
\label{Sec:appcasecompdisentangle}
In this section, we provide more details and more examples about the semantic contexts that maximally trigger these in 
reasoning components (\cref{subsec: what}).
We aggregate the textual contexts across the GSM8K dataset that yield the 
highest activations for our top-ranked, IPG-identified features.
We include the results in~\cref{fig:sae_features_20,fig:sae_features_24,fig:sae_features_26}.
\paragraph{Experiment Setup: }
We analyze Sparse Autoencoder (SAE) features extracted from the Qwen2.5‑Math‑1.5B‑Instruct model. 
For each feature index of interest (e.g., those identified by IPG from a given layer), we obtain the total activation counts, peak frequency tokens, and the top‑activation values along with their corresponding token contexts (a window of 17 tokens centered on the activation peak). 
This allows us to qualitatively inspect the behavior of individual SAE features. The experiments are performed on the GSM8K task, focusing on layer \{20,24,26\}.

\subsubsection{Generated Responses before and after Intervention}
\label{Sec:appcasegenerateresponses}
In this section, we provide more examples with our \method framework corresponding to~\cref{Sec:intervene}.
We present generated responses before and after intervention on multiple models, spanning different reasoning benchmarks and intervention granularities (neuron-level vs. SAE feature-level). Across models and tasks, we can consistently observe that steering IPG-identified components leads to more coherent intermediate reasoning steps and improved final answers.

\newpage
\begin{tcolorbox}[colback=orange!10, colframe=brown!80!black, title=\textbf{Question}]
How many r are there in the word strawberry? 
\end{tcolorbox}

\begin{tcolorbox}[colback=blue!5, colframe=black, title=\textbf{Response}]

\textbf{Original:}
\begin{DiffVerb}
To solve this problem, I need to count the number of letters in the word “strawberry.” Okay, so first, I'll start by writing down the word: strawberry. Let me see, that's an 8-letter word. Hmm, wait, let me count each letter to be sure.
Starting from the beginning: s, t, r, a, w, b, r, e, r. So that‘s s (1), t (2), r (3), a (4), w (5), b (6), r (7), e (8), r (9). Wait, that’s nine letters. But I thought it was an 8-letter word. Maybe I‘m miscounting. Wait, let me check again. Maybe I missed a letter. So, “strawberry” is spelled S-T-R-A-W-B-E-R-R-Y. Wait, that’s 9 letters. Hmm, maybe I‘m getting confused. Let me write it out: S (1), T (2), R (3), A (4), W (5), B (6), E (7), R (8), R (9), Y (10). Wait, that‘s 10 letters. Now I’m really confused. Is it 10 letters? Or is it 9? Wait, let me look it up. Wait, no, I can‘t look it up, I need to count. Okay, let’s try again. S-T-R-A-W-B-E-R-R-Y. So S is 1, T is 2, R is 3, A is 4, W is 5, B is 6, E is 7, R is 8, R is 9, Y is 10. So it‘s 10 letters. But I thought it was an 8-letter word. Wait, maybe it’s a different word. Wait, maybe the word is “strawberry” without the ‘y’? Let me check. No, the spelling is S-T-R-A-W-B-E-R-R-Y, which is 10 letters. So the problem must be about the letter ‘r’ instead of the word. So the question is, how many ‘r’s are there in the word “strawberry.” So if the word is 10 letters, how many ‘r’s are there? Wait, let me count again. S-T-R-A-W-B-E-R-R-Y. !R{So the letters are: 1: S 2: T 3: R 4: A 5: W 6: B 7: E 8: R 9: R 10: Y. So there are two ‘R’s: one at position 3 and another at position 8. So the answer should be 2. 
</think>
2}
\end{DiffVerb}
\vspace{0.5cm}
\hrule
\vspace{0.5cm}

\textbf{Enhanced:}
\begin{DiffVerb}
Okay, so I need to figure out how many "r"s are in the word "strawberry." Let me start by writing down the word: S-T-R-A-W-B-E-R-R-Y. Hmm, that's a bit long, so I should count each "r" individually. 
First, I notice the letters S, T, R, A, W, B, E, R, R, Y. Let's break them down one by one. The first R is after S and T, so that's one "r." Then, moving along, the next R is right before E, so that's another "r." After that R, there are two more R's before Y. So let me count them:
1. R after S 2. R after T 3. R after A 4. R after B 5. R after E 6. R before Y
!G{Wait, that‘s six R’s. Let me make sure I didn‘t miss any.} Let me go through each letter again: 1. S: Not an R. 2. T: Not an R. 3. R: 1 4. A: Not an R. 5. W: Not an R. 6. B: Not an R. 7. E: Not an R. 8. R: 2 9. R: 3 10. Y: Not an R.
So yeah, only three R's. Wait, that contradicts my previous count. Hmm, maybe I miscounted earlier. Let me recount more carefully.
Let me list each letter and note if it's an R:
1. S: No 2. T: No 3. R: Yes (1) 4. A: No 5. W: No 6. B: No 7. E: No 8. R: Yes (2) 9. R: Yes (3) 10. Y: No
!G{Ah, so only three R's. So the answer should be 3. I think I confused myself earlier, but now it's clear. So the total number of "r"s in "strawberry" is 3.
</think>
3}

\end{DiffVerb}

\end{tcolorbox}
Case Study on Deepseek-R1-Distilled-Qwen1.5B . Steering one neuron with highest IPG score with scaling factor $\gamma=1.1$ on layer 20.

\newpage
\begin{tcolorbox}[colback=orange!10, colframe=brown!80!black, title=\textbf{Question}]
Ellen decided to play a prank on her friend. She got a case of 12 sodas and shook 3 of them up. Then she took 1 unshaken soda for herself and left. Ellen's brother stopped by and took 1 of the shaken sodas and 2 of the unshaken sodas, then Ellen's friend came along. What is the likelihood, expressed as a percentage, that Ellen's friend gets sprayed with soda from a shaken can?
\end{tcolorbox}

\begin{tcolorbox}[colback=blue!5, colframe=black, title=\textbf{Response}]

\textbf{Original:}
\begin{DiffVerb}
To determine the likelihood that Ellen‘s friend gets sprayed with soda from a shake can, we need to follow the sequence of events and calculate the number of sodas in each state.\n\n1. **Initial number of sodas:**\n Ellen starts with 12 sodas.\n\n!R{2. **Sodas after Ellen shakes 3:**\n Ellen shakes 3 sodas, so there are \\(12 - 3 = 9\\) sodas left.\n\n}3. **Sodas after Ellen takes 1 unshaken soda:**\n Ellen takes 1 unshaken soda, so there are \\(9 - 1 = 8\\) sodas left, with 3 sodas still shaken.\n\n4. **Sodas after Ellen’s brother takes 1 shake and 2 unshaken sodas:**\n Ellen‘s brother takes 1 shake soda and 2 unshaken sodas, so there are \\(8 - 1 - 2 = 5\\) sodas left, with \\(3 - 1 = 2\\) sodas still shaken.\n\n5. **Sodas after Ellen’s friend comes along:**\n !R{Ellen‘s friend now has the remaining 5 sodas, with 2 of them being shake sodas.}\n\nThe probability that Ellen’s friend gets sprayed with soda from a shake can is the ratio of the number of shake sodas to the total number of sodas Ellen‘s friend has. This probability is calculated as follows:\n\\[\n\\text{Probability} = \\frac{\\text{Number of shake sodas}}{\\text{Total number of sodas}} = \\frac{2}{5}\n\\]\n\nTo express this probability as a percentage, we multiply by 100:\n\\[\n\\frac{2}{5} \\times 100 = 40\\%\n\\]\n\n 
!R{Thus, the likelihood that Ellen's friend gets sprayed with soda from a shake can is \\(\\boxed{40}\\).}


\end{DiffVerb}

\vspace{0.5cm}
\hrule
\vspace{0.5cm}

\textbf{Enhanced:}
\begin{DiffVerb}
To determine the likelihood that Ellen‘s friend gets sprayed with soda from a shake can, we need to follow the sequence of events and calculate the number of sodas in each state.\n\n1. **Initial number of sodas:**\n Ellen starts with 12 sodas.\n\n!G{2. **Sodas after Ellen shakes 3:**\n Ellen shakes 3 sodas, so she has 9 sodas left unshaken.\n \\[\n \\text{Unshaken sodas} = 12 - 3 = 9\n \\]\n \\[\n \\text{Shaken sodas} = 3\n \\]\n\n}3. **Sodas after Ellen takes 1 unshaken soda:**\n Ellen takes 1 unshaken soda for herself, so she has 8 unshaken sodas left.\n \\[\n \\text{Unshaken sodas} = 9 - 1 = 8\n \\]\n \\[\n \\text{Shaken sodas} = 3\n \\]\n\n4. **Sodas after Ellen’s brother takes 1 shake and 2 unshake:**\n Ellen‘s brother takes 1 shake soda and 2 unshake sodas, so the number of sodas changes as follows:\n \\[\n \\text{Shake sodas} = 3 - 1 = 2\n \\]\n \\[\n \\text{Unshake sodas} = 8 - 2 = 6\n \\]\n\n5. **Sodas after Ellen’s friend comes along:**\n Ellen‘s friend comes along and gets 1 soda. The probability that this soda is from a shake can is the number of shake sodas divided by the total number of sodas left.\n \\[\n \\text{Total sodas left} = 2 + 6 = 8\n \\]\n \\[\n \\text{Probability that Ellen’s friend gets a shake soda} = \\frac{2}{8} = \\frac{1}{4} = 0.25\n \\]\n \\[\n \\text{Percentage} = 0.25 \\times 100 = 25 \% \n \\]\n\n !G{Therefore, the likelihood that Ellen's friend gets sprayed with soda from a shake can is \\(\\boxed{25}\\).}
\end{DiffVerb}
\end{tcolorbox}

Case Study on GSM8K for Qwen2.5-Math-1.5B-Instruct. Steering 5\% \method neurons with scaling factor $\gamma=1.1$ on layer 20.

\newpage
\begin{tcolorbox}[colback=orange!10, colframe=brown!80!black, title=\textbf{Question}]
John wins an award at work. The award has a 1 time monetary reward of \$5000 and a 5\% raise in salary. If he makes 2000 a week, every week of the year and he got the award on the first day of the year how much money did he make that year?
\end{tcolorbox}

\begin{tcolorbox}[colback=blue!5, colframe=black, title=\textbf{Response}]

\textbf{Original:}
\begin{DiffVerb}
To determine how much money John made that year, we need to consider both his regular weekly earnings and the additional monetary reward he received from the award.\n\n1. **Calculate John‘s regular annual earnings:**\n John makes $2000 per week. There are 52 weeks in a year.\n \\[\n \\text{Regular annual earnings} = 2000 \\times 52 = 104000\n \\]\n\n2. **Calculate the monetary reward from the award:**\n The award has a 1 time monetary reward of $5000.\n\n3. **Calculate the total earnings for the year:**\n Add the regular annual earnings to the monetary reward.\n \\[\n \\text{Total earnings} = 104000 + 5000 = 109000\n \\]\n\n
!R{Therefore, the total amount of money John made that year is \\(\\boxed{109000}\\).}
\end{DiffVerb}

\vspace{0.5cm}
\hrule
\vspace{0.5cm}

\textbf{Enhanced:}
\begin{DiffVerb}
To determine how much money John made that year, we need to consider both his regular weekly earnings and the additional monetary reward he received from the award.\n\n1. **Calculate John's regular annual earnings:**\n John makes \$2000 per week. There are 52 weeks in a year.\n \\[\n \\text{Regular annual earnings} = 2000 \\times 52 = 104000\n \\]\n\n2. **Calculate the monetary reward from the award:**\n The award has a 1 time monetary reward of \$5000.\n\n!G{3. **Calculate the 5\% raise in salary:**\n The 5\% raise in salary is applied to his regular weekly earnings.\n \\[\n \\text{Raise per week} = 2000 \\times 0.05 = 100\n \\]\n Since the raise is applied to each week, the total raise for the year is:\n \\[\n \\text{Total raise for the year} = 100 \\times 52 = 5200\n \\]\n\n}4. **Calculate the total earnings for the year:**\n Add the regular annual earnings, the monetary reward, and the total raise.\n \\[\n \\text{Total earnings} = 104000 + 5000 + 5200 = 114200\n \\]\n
\G{Therefore, the total amount of money John made that year is \\(\\boxed{114200}\\)."}
\end{DiffVerb}

\end{tcolorbox}
Case Study on GSM8K for Qwen2.5-Math-1.5B-Instruct. Steering one \method SAE feature with scaling factor $\gamma=1.8$ on layer 20.

\newpage
\begin{tcolorbox}[colback=orange!10, colframe=brown!80!black, title=\textbf{Question}]
Cars have lined up on the motorway. Some of the cars drive through in the first 15 minutes of the traffic jam, then 20 more cars drive through in the remaining 15 minutes of the jam. 5 cars from the line take an exit so they don't have to drive through the traffic jam. If there were originally 30 cars on the motorway, how many cars drove through the traffic jam in the first 15 minutes?
\end{tcolorbox}

\begin{tcolorbox}[colback=blue!5, colframe=black, title=\textbf{Response}]

\textbf{Original:}
\begin{DiffVerb}
To find the number of cars that drove through the traffic jam in the first 15 minutes, we need to first determine the total number of cars that drove through the traffic jam.\n\nWe know that 20 cars drove through in the remaining 15 minutes. \n\nLet‘s denote the number of cars that drove through in the first 15 minutes as ’x‘. \n\took an exit, we need to subtract these from the total number of n. !R{Since 5 cars cars to find the number of cars that actually drove through the traffic jam. \n\nSo, the total number of cars that drove through the traffic jam is (x + 20) - 5.\n\nWe are given that there were originally 30 cars on the motorway. \n\nTherefore, (x + 20) - 5 = 30.\n\n} Simplifying the equation, we get x + 20 - 5 = 30.\n\nx + 15 = 30.\n\n !R{Subtracting 15 from both sides, we get x = 15.\n\nSo, the number of cars that drove through the traffic jam in the first 15 minutes is 15.\n\n\\boxed{15}}
\end{DiffVerb}

\vspace{0.5cm}
\hrule
\vspace{0.5cm}

\textbf{Enhanced:}
\begin{DiffVerb}
To find the number of cars that drove through the traffic jam in the first 15 minutes, we need to first determine the total number of cars that drove through the traffic jam.\n\nWe know that 20 cars drove through in the remaining 15 minutes. \n\nLet's denote the number of cars that drove through in the first 15 minutes as 'x'. \n\n!G{Since 5 cars took an exit, the total number of cars that drove through the traffic jam is the original number of cars minus the cars that took an exit, which is 30 - 5 = 25.\n\n}We also know that the total number of cars that drove through the traffic jam is the sum of the cars that drove through in the first 15 minutes and the cars that drove through in the remaining 15 minutes. \n\n!G{So, we can write the equation: x + 20 = 25.\n\nTo solve for 'x', we subtract 20 from both sides of the equation: x = 25 - 20.\n\nx = 5.\n\n}However, this is the number of cars that drove through the traffic jam in the first 15 minutes, but it does not account for the fact that the question asks for the number of cars that drove through the traffic jam in the first 15 minutes, not the number of cars that were in the first 15 minutes. \n\nSince 5 cars took an exit, the number of cars that were in the first 15 minutes is 30 - 5 = 25. \n\nHowever, we know that 20 cars drove through in the remaining 15 minutes. \n\n!G{So, the number of cars that drove through in the first 15 minutes is 25 - 20 = 5.}
\end{DiffVerb}

\end{tcolorbox}
Case Study on GSM8K for Llama-3.1-8B-Instruct. Steering 5\% IPG neurons with scaling factor $\gamma=1.1$ on layer 22.

\newpage
\begin{tcolorbox}[colback=orange!10, colframe=brown!80!black, title=\textbf{Question}]
A Senate committee has 5 Democrats, 5 Republicans, and 1 Independent. In how many ways can they sit around a circular table if all the members of each party all sit next to each other? (Two seatings are considered equivalent if one is a rotation of the other.)

\end{tcolorbox}
\begin{tcolorbox}[colback=blue!5, colframe=black, title=\textbf{Response}]

\textbf{Original:}
\begin{DiffVerb}
To solve this problem, we need to consider the restrictions and the total number of arrangements.\n\n1. **Total number of people**: There are 11 people in total (5 Democrats, 5 Republicans, and 1 Independent).\n\n2. **Restrictions**: All members of each party must sit next to each other. This means we can treat each party as a single unit when arranging the people around the table.\n\n3. **Circular table**: Since the table is circular, we need to account for the fact that rotations of the same arrangement are considered equivalent. This means we will have to divide the total number of arrangements by the number of people (11) to avoid overcounting.\n\n!R{4. **Arranging the parties**: There are 3 parties (Democrats, Republicans, and Independent), which can be arranged in 3! = 6 ways around the table.\n\n}5. **Arranging the members within each party**: Within each party, the members can be arranged in the following ways:\n - Democrats: 5! ways\n - Republicans: 5! ways\n - Independent: 1 way (since there is only 1 member)\n\n!R{6. **Total number of arrangements**: The total number of arrangements is the product of the number of ways to arrange the parties and the number of ways to arrange the members within each party:\n - Total = 6 * 5! * 5! * 1\n\n7. **Accounting for rotations**: Since the table is circular, we need to divide the total number of arrangements by the number of people (11) to avoid overcounting:\n - Total = (6 * 5! * 5! * 1) / 11\n\n8. **Calculating the total**: Now, let's calculate the total number of arrangements:\n - 5! = 5 * 4 * 3 * 2 * 1 = 120\n - Total = (6 * 120 * 120 * 1) / 11\n - Total = (6 * 14400) / 11\n - Total = 86400 / 11\n - Total = 7872\n\nSo, the total number of ways the Senate committee can sit around a circular table with all members of each party sitting next to each other is:\n\n\\boxed{7872}”.}
\end{DiffVerb}

\hrule

\textbf{Enhanced:}
\begin{DiffVerb}
To solve this problem, we need to consider the restrictions and the total number of arrangements.\n\n1. The members of each party must sit together. This means we can treat each party as a single unit. So, we have 3 units: Democrats, Republicans, and the Independent.\n\n2. Since the table is circular, we need to account for the fact that rotations of the same arrangement are considered equivalent. This means we will have to divide the total number of arrangements by the number of people (or units) in each arrangement.\n\n!G{3. First, let‘s calculate the number of ways to arrange these 3 units around the circular table. There are (3-1)! = 2! = 2 ways to arrange the units.\n\n4. Now, let’s consider the arrangements within each party. There are 5! ways to arrange the Democrats, 5! ways to arrange the Republicans, and 1 way to arrange the Independent (since there is only one Independent).\n\n5. To find the total number of arrangements, we multiply the number of ways to arrange the units by the number of ways to arrange the members within each party: 2 * 5! * 5! * 1.\n\n6. Now, we calculate the total number of arrangements: 2 * 5! * 5! = 2 * 120 * 120 = 28800.\n\n7. Therefore, the total number of ways the Senate committee can sit around a circular table if all the members of each party all sit next to each other is 28800.\n\n\\boxed{28800}.}
\end{DiffVerb}

\end{tcolorbox}
Case Study on Math500 for Llama-3.1-8B-Instruct. Steering 5\% IPG neurons with scaling factor $\gamma=1.1$ on layer 22.

\subsection{Comparison with Training-based Behavior Control for Long-Horizon Reasoning}

\label{Sec::appReFTcomp}
\paragraph{Experimental Setup.} To demonstrate our \method is good at handling long-term dependencies, we additionally adopt ReFT \citep{wu2024reft} as a baseline. ReFT is a powerful training-based method for controlling model behavior by optimizing a language modeling loss on representation space. Specifically, to ensure a fair comparison, we trained ReFT on the same single layer (Layer 22 for Qwen model, Layer 24 for Llama model) using 1,000 samples from GSM8K train split. We used the best hyperparameter configuration for ReFT. We adopt the same evaluation settings in our paper (greedy decoding, zero-shot on GSM8K test set) and report only enhancement results as ReFT does not support suppression experiment.

\paragraph{Results and Analysis.} 
As shown in Table \ref{tab:ipg_reft_comparison}, IPG consistently outperforms ReFT on both models under our setting. Notably, ReFT leads to a performance drop under our setting. We hypothesize three reasons for this degradation. (i) ReFT loss optimizes language modeling, not reasoning. (ii) Overfitting on training data. (iii) Data inefficiency of training-based method. Based on the empirical results, we confirm the advantages of \method over handling long-term dependencies in multi-step reasoning scenario.

\begin{table}[t]
    \centering
    \caption{Performance comparison across models under different methods.}
    \begin{tabular}{lccc}
        \toprule
        \textbf{Models} & \textbf{Original} & \textbf{IPG (Enhanced)} & \textbf{ReFT} \\
        \midrule
        Qwen2.5-Math1.5B-Instruct & 82.41 & \textbf{84.76} & 62.47 \\
        Llama3.1-8B-Instruct     & 85.89 & \textbf{87.41} & 77.94 \\
        \bottomrule
    \end{tabular}
    \label{tab:ipg_reft_comparison}
\end{table}

\subsection{Behavior-control beyond Reasoning}
\label{Sec:appotheraspects}
\subsubsection{Reflection Behavior}

\paragraph{Experimental Setup.} To demonstrate flexibility of the signal function of $J$, we instantiate it to capture \textit{reflection}. We utilize a set of typical keywords associated with reflective behavior serving as the signal (e.g., ``wait'', ``but'', ``double check''). 
For the signal function, we calculate the total number of times these words appear in the trajectory $\tau$, while preserving their token position information.

\paragraph{Results and Analysis.} We conduct experiments using DeepSeek-Qwen-1.5B~\citep{guo2025deepseek} on the AIME 2024 benchmark~\citep{aime2024}. 
As shown in Table~\ref{tab:reflection_results}, our intervention significantly increases the average number of reflection tokens per sample from 113.9 to 156.5. 
Consequently, this enhanced reflective behavior drives an improvement in final task accuracy, rising from 16.67\% to 20.00\%. 
This result confirms that our framework can effectively amplify specific mechanistic behaviors to improve complex reasoning outcomes.

\begin{table}[ht]
    \centering
    \caption{Impact of intervening on reflection-based signals on AIME 2024. We report the task accuracy and the average number of reflection tokens per sample (calculated over 30 samples).}
    \label{tab:reflection_results}
    \begin{tabular}{lcc}
        \toprule
        \textbf{Method} & \textbf{Accuracy (\%)} $\uparrow$ & \textbf{Reflection Tokens (Avg.)} $\uparrow$ \\
        \midrule
        Original & 16.67 & 113.9 \\
        \textbf{Intervened (Ours)} & \textbf{20.00} & \textbf{156.5} \\
        \bottomrule
    \end{tabular}
\end{table}
\subsubsection{Safety}
\paragraph{Experimental Setup.}
Additionally, we try to attribute and control safety awareness. We instantiate the signal function $J$ as a binary reward with regard to safety. For each prompt, we generate model responses both \emph{before} and \emph{after} intervention. Each response is then classified as \emph{harmful} or \emph{non-harmful} using the \textsc{WildGuard} classifier~\citep{han2024wildguard}. We report the \emph{harmlessness rate}, defined as
\[
\text{Harmlessness Rate} = \frac{\#\text{non-harmful responses}}{\#\text{total responses}}.
\]
A higher harmlessness rate indicates safer model behavior. As a baseline, we adopt the activation-based method used in our main experiments, which selects neurons with the highest activation values. Following the main paper, we use greedy decoding for all generations. The \textsc{Enhance} setting amplifies components associated with reduced harmfulness (higher is better), while \textsc{Suppress} attenuates these components (lower is better).

\paragraph{Results and Analysis.}
Table~\ref{tab:wildguard} reports the results on Qwen2.5-1.5B-Instruct. IPG consistently outperforms the activation-based baseline under both enhancement and suppression settings. These results suggest that IPG is able to identify internal components that causally contribute to harmful behavior, enabling more effective and controllable safety intervention.

\begin{table}[h]
\centering
\caption{Comparison of IPG and activation-based methods for controlling harmfulness on Qwen2.5-1.5B-Instruct using the \textsc{WildGuard} benchmark~\citep{han2024wildguard}.}
\label{tab:wildguard}
\begin{tabular}{lccc}
\toprule
Method & Original (\%) & Enhance $\uparrow$ (\%) & Suppress $\downarrow$ (\%) \\
\midrule
Activation & 89.52 & 89.64 & 89.05 \\
IPG (Ours) & 89.52 & \textbf{90.27} & \textbf{85.81} \\
\bottomrule
\end{tabular}
\end{table}
\subsection{Scalability of \method}
\label{sec:appscalability}
Beyond our primary evaluations on architectures (e.g., Qwen2.5-Math-1.5B-Instruct and Llama-3-8B-Instruct), we investigate the scalability of \method by applying it to a larger-scale model, Qwen3-14B.
The objective is to verify whether our framework can consistently localize and modulate reasoning-critical components as model capacity and architectural complexity increase.

We apply \method to identify and intervene on crucial neuron-level components within Qwen3-14B.
Using standard greedy decoding, we evaluate our \method-enhanced model across three challenging reasoning benchmarks: GPQA, AIME, and MATH500. 

\begin{table}[htbp]
    \centering
    \small
    \caption{Performance comparison of the baseline Qwen3-14B and the \method-enhanced model (intervening on neurons) using greedy decoding. Metrics are reported as absolute accuracy (\%).}
    \label{tab:scalability}
    \begin{tabular}{lccc}
        \toprule
        \textbf{Model Setup} & \textbf{GPQA} & \textbf{AIME} & \textbf{MATH500} \\
        \midrule
        Qwen3-14B (Original) & 27.27 & 33.33 & 73.60 \\
        Qwen3-14B (\method)  & \textbf{34.34} & \textbf{40.00} & \textbf{76.00} \\
        \midrule
        \textit{$\Delta$ Improvement} & \textit{+7.07} & \textit{+6.67} & \textit{+2.40} \\
        \bottomrule
    \end{tabular}
\end{table}

As demonstrated in \cref{tab:scalability}, \method yields consistent performance improvements across all evaluated datasets.
These enhancements validate that \method is not constrained by model size. 
Regardless of scale, our framework robustly pinpoints the output-oriented mechanisms driving advanced reasoning, enabling reliable, granular capability modulation even in highly parameterized models.

\section{Use of Large Language Models (LLMs)}
\label{sec:appusellm}
We use large language models (LLMs) as auxiliary tools in the writing process, specifically to help identify grammatical errors and polish our writing.
Beyond this, LLMs were not involved in core research contributions, including the design of methods, theoretical analysis, and experimental setup. All analyses, and conclusions presented remain the sole responsibility of the authors.
\section{Broader Impact}
\label{sec:appborderimpact}

We believe there are benefits of advancing interpretable, controllable reasoning,
especially given that our method
improves existing reasoning models. 
Our method can have clear societal benefits: more efficient and controllable AI assistants could broaden access to high‑quality educational support,
scientific reasoning, and problem‑solving tools.

\end{document}